\documentclass[10pt,twocolumn,letterpaper,table]{article}

\usepackage{iccv}              

\usepackage{microtype}

\definecolor{iccvblue}{rgb}{0.21,0.49,0.74}
\usepackage[pagebackref,breaklinks,colorlinks,allcolors=iccvblue]{hyperref}
\usepackage{xcolor}
\usepackage{graphicx}
\usepackage{booktabs}
\usepackage{nicefrac}
\usepackage{blkarray, bigstrut}
\usepackage{amsmath}
\usepackage{amssymb}
\usepackage{mathtools}
\usepackage{amsthm}
\usepackage{array}
\usepackage{multirow}
\DeclareMathOperator*{\argmax}{arg\,max}

\usepackage[capitalize,noabbrev]{cleveref}
\usepackage{enumitem}
\usepackage{subcaption}
\usepackage{caption}
\usepackage{tikz}
\usetikzlibrary{bayesnet, calc, arrows.meta, bending}
\usetikzlibrary{decorations.pathreplacing,angles,quotes,shapes.multipart, calligraphy}

\usepackage{minitoc}
\usepackage[toc,page,header]{appendix}

\newcommand{\changelinkcolor}[1]{\hypersetup{linkcolor=#1}}  
\AtBeginDocument{%
  
}

\theoremstyle{plain}
\newtheorem{theorem}{Theorem}[section]
\newtheorem{proposition}[theorem]{Proposition}
\newtheorem{lemma}[theorem]{Lemma}

\theoremstyle{definition}
\newtheorem{definition}[theorem]{Definition}

\theoremstyle{remark}
\newtheorem{remark}[theorem]{Remark}

\def\paperID{14208} 
\def\confName{ICCV}
\def\confYear{2025}

\title{Flow Stochastic Segmentation Networks}

\newcommand{\cand}{\hspace{0.7em}}

\author{Fabio De Sousa Ribeiro\footnotemark[2] \cand Omar Todd \cand Charles Jones \cand Avinash Kori \cand Raghav Mehta \cand Ben Glocker\footnotemark[2]
\\[3pt]
Imperial College London
}

\begin{document}
\doparttoc
\faketableofcontents%
\maketitle
\renewcommand*{\thefootnote}{\fnsymbol{footnote}}
\begin{abstract}
    We introduce the Flow Stochastic Segmentation Network (Flow-SSN), a generative segmentation model family featuring discrete-time autoregressive and modern continuous-time flow variants. We prove fundamental limitations of the low-rank parameterisation of previous methods and show that Flow-SSNs can estimate arbitrarily high-rank pixel-wise covariances without assuming the rank or storing the distributional parameters. Flow-SSNs are also more efficient to sample from than standard diffusion-based segmentation models, thanks to most of the model capacity being allocated to learning the base distribution of the flow, constituting an expressive prior. We apply Flow-SSNs to challenging medical imaging benchmarks and achieve state-of-the-art results. 
    Code available: \textup{\url{https://github.com/biomedia-mira/flow-ssn}}.
    \footnotetext[2]{Email: {\{f.de-sousa-ribeiro, bglocker\}@imperial.ac.uk}
    }
\end{abstract}
\renewcommand*{\thefootnote}{\arabic{footnote}}

\section{Introduction}
\label{sec: introduction}
Semantic image segmentation consists of producing pixel-wise predictions that reflect objects and their boundaries. Traditional methods typically approach this as a deterministic pixel-wise classification task, while overlooking the inherent \textit{uncertainty} of the spatial structures involved~\citep{zheng2015conditional,chen2017deeplab,kendall2017uncertainties}. Uncertainty generally relates to unknown or imperfect information,
often arising from a lack of knowledge, partial observations, and/or inherently stochastic events. Uncertainty is commonly decomposed into two distinct parts~\citep{der2009aleatory,kendall2017uncertainties}: {(i)} \textit{Epistemic} uncertainty, which relates to a lack of knowledge, and it can be reduced in principle by observing more data; {(ii)} \textit{Aleatoric} uncertainty, which relates to inherent unknowns that differ each time we run the same experiment (e.g. flipping a coin), and it cannot be reduced by observing more data. Although conceptually attractive, the utility and validity of this decomposition is actively contested~\citep{smith2024rethinking,kirchhof2025position}.

Inherent ambiguities are particularly prevalent in medical imaging, where medical opinions can vary significantly across different experts~\citep{kohl2018probabilistic, joskowicz2019inter,bran2024qubiq,barfoot2024average}. In this context, uncertainty can arise from several factors, including indistinct boundaries, occlusion, poor image acquisition quality, and the intrinsic variability of the underlying pathology. Therefore, to effectively model these ambiguities and reflect the real-world variability of expert opinions, segmentation models ought to capture a rich \textit{distribution} of plausible segmentation outcomes~\citep{kohl2018probabilistic,czolbe2021segmentation}. Furthermore, providing uncertainty estimates is useful for revealing whether a model has sufficient knowledge to provide a reliable assessment, which is important in safety-critical real-world settings~\cite{bernhardt2022failure,pmlr-v235-papamarkou24b,li2025position}.

There is a growing interest in using probabilistic methods for estimating uncertainty in image segmentation. Most existing methods handle uncertainty by factorising the output posterior into per-pixel marginal distributions, thereby ignoring any correlations between pixels. Therefore, pixel-wise independent uncertainty estimates are incapable of fully capturing spatially structured uncertainty~\citep{nehme2023uncertainty,langley2022structured}. To address this, \citet{monteiro2020stochastic} proposed Stochastic Segmentation Networks (SSNs), which can explicitly model spatially correlated aleatoric uncertainty without requiring variational approximations or latent variable assumptions. Their method involves placing a low-rank multivariate
Gaussian distribution over the logit space (i.e. before the softmax), then using Monte Carlo integration to marginalise out the logits and compute pixel-wise joint likelihoods~\citep{kendall2017uncertainties,kendall2018multi}. 
Although promising, the trouble with SSNs is three-fold: (i) The assumed rank of the low-rank approximation is typically kept small due to computational constraints (e.g., $\simeq$10), and it is almost surely underspecified relative to the true rank of high-dimensional, pixel-wise covariances; (ii) They often require an expensive mean pre-training stage to ensure proper convergence, as jointly optimising poor initial estimates of the mean and covariance can lead to getting trapped in suboptimal minima; (iii) They suffer from training instabilities, partly due to a lack of guarantee that the low-rank covariance matrix remains positive definite throughout training.

\noindent\textbf{Contributions.} We propose Flow Stochastic Segmentation Networks (Flow-SSNs), a generative segmentation model class featuring discrete-time autoregressive and modern continuous-time flow parameterisations. Flow-SSNs can estimate arbitrarily high-rank pixel-wise covariances without assuming the rank a priori, storing distributional parameters, or assuming a lower-dimensional latent space as in VAEs. Flow-SSNs are more efficient to sample from than typical diffusion-based segmentation models, as most of the model capacity is dedicated to learning a flow's \textit{prior}, while the flow itself is lightweight. In summary, our contributions are:

\begin{enumerate}[leftmargin=20pt]
    \item[\S\ref{sec: motivation}] We prove 
    fundamental limitations of the low-rank parameterisation of SSNs by showing that the effective rank grows sublinearly with the assumed rank;
    
    \item[\S\ref{sec: flow ssn}] We introduce Flow-SSNs, a generative segmentation model capable of modelling complex covariance structures efficiently by learning a flow's \textit{prior} and using a lightweight flow to model pixel-wise dependencies;
    
    \item[\S\ref{sec: experiments}] Applying Flow-SSN to a toy problem and two real-world medical image segmentation benchmarks, we show state-of-the-art results with fewer parameters.
\end{enumerate}
\section{Related Work}
\label{sec: related work}
Existing work on stochastic segmentation can be broadly categorised as: {(i)} Bayesian methods which approximate a posterior over neural network parameters~\citep{kendall2017uncertainties}; {(ii)} latent variable generative models~\citep{kohl2018probabilistic,baumgartner2019phiseg,wu2024medsegdiff}; and {(iii)} distributional/evidential methods, which estimate a complex joint distribution directly in pixel space~\citep{malinin2018predictive,sensoy2018evidential,monteiro2020stochastic,ulmer2023prior}.~\citet{mackay1992bayesian},~\citet{neal1993probabilistic} and \citet{hinton1993keeping} laid the foundations for modern-day Bayesian Neural Networks (BNNs), which approximate neural network parameter posteriors and enable uncertainty estimation. More recently,~\citet{kendall2017uncertainties,kwon2020uncertainty}, and others~\cite{kendall2018multi,de2020deep,de2020introducing,yang2022st++}, used these techniques for classification/segmentation, but handled uncertainty by factorising the output posterior into per-pixel marginal distributions, thereby ignoring pixel correlations.

Stochastic segmentation methods based on the Variational Autoencoder (VAE)~\citep{kingma2013auto,rezende2014stochastic} framework implicitly assume that the data resides in a lower-dimensional latent manifold and hope that the pixel-wise independent decoder will learn to translate uncorrelated latent variables into meaningful spatial variation in pixel space~\citep{kohl2018probabilistic,baumgartner2019phiseg,zepf2023label}.~\citet{selvan2020uncertainty,valiuddin2021improving} also use a VAE for segmentation but apply a normalising flow~\citep{tabak2013family} to the latent variables to make them more expressive. Since providing latent variable identifiability guarantees is challenging for most problems~\citep{hyvarinen2024identifiability}, one often resorts to unfalsifiable assumptions about both the functional form and dimensionality of the latent space. Although VAEs can work well for certain segmentation tasks~\citep{kohl2018probabilistic,mehta2025cf}, they sometimes underperform in high-dimensional settings, and are subject to the \textit{prior hole} problem~\citep{rezende2018taming,lucas2019don,Ghosh2020From,ribeiro2025demystifying}.

Diffusion models~\citep{sohl2015deep,ho2020denoising} are a promising viable alternative which has been recently explored for producing stochastic segmentations~\citep{amit2021segdiff,rahman2023ambiguous,zbinden2023stochastic,wu2024medsegdiff}. However, their high inference costs can restrict their usability for medical experts to edit segmentation annotations in real-time. Recent work on Continuous Normalising Flows (CNFs)~\citep{chen2018neural} provides efficient (simulation-free) ways to learn straighter paths between distributions compared to diffusion paths ~\citep{lipman2023flow,liu2023flow,albergo2023building,tong2024improving}. CNFs that induce straighter paths are computationally more efficient to solve, making them an attractive option for real-time editing of medical imaging annotations. There is limited prior work on exploring CNFs for segmentation;~\cite{bogensperger2024flowsdf} combine Flow Matching (FM)~\citep{lipman2023flow} with the signed distance function (SDF) for image segmentation, but their investigation is restricted to binary data. Our approach differs substantially in its formulation as it is defined within an SSN-like~\citep{monteiro2020stochastic} paradigm, and naturally extends to categorical data. Specifically, our model generates multiple segmentations by sampling from an expressive, learned base distribution (i.e. prior) conditional on the image, rather than random noise. 

Limited prior work exists on using autoregressive models for stochastic segmentation tasks.~\citet{zhang2022pixelseg} proposed an autoregressive approach using a PixelCNN~\citep{van2016conditional,salimans2017pixelcnn}, which can learn full rank pixel-wise covariances. However, it requires each pixel to be generated sequentially, which is slow. To mitigate this, they use a downsampled resolution, discarding input information. 
SSNs~\citep{monteiro2020stochastic} provide a simpler alternative for learning the joint distribution over pixel-wise label maps that does not require making latent variable assumptions or variational approximations. Multiple works build on SSNs to enable fine-grained sample control~\citep{nussbaum2022structuring}, learning mixtures of stochastic experts~\citep{gao2022modeling} and conditioning on label style~\citep{zepf2023label}. However, SSNs are subject to training instabilities and make strong assumptions about the rank of the true pixel-wise covariance being quite small, which is under-determined for most problems. Flow-SSNs make no such assumptions and can estimate arbitrarily high-rank covariances. Concurrently,~\cite{tschannen2025jetformer,zhai2025normalizing,gu2025starflow} revisit classical flow models~\citep{dinh2017density,papamakarios2017masked,kingma2018glow} for generation, but not for segmentation.
\section{Preliminaries}
\label{sec: background}
Let $\{(\mathbf{x}_i, \mathbf{y}_i)\}_{i=1}^n$ be a dataset of $n$ image $\mathbf{x}_i \in \mathbb{R}^{c\times d}$ and one-hot label map $\mathbf{y}_i \in \{0,1\}^{k\times d}$ pairs, with number of channels $c$, height times width $hw = d$, and categories $k$. We denote $\mathrm{softmax}_k(\cdot)$ as the row-wise softmax applied after reshaping the input from $\mathbb{R}^{kd}$ to $\mathbb{R}^{k\times d}$. To avoid cluttered notation, we may use lowercase symbols to denote both random variables and their realisations when context permits. 
\\[5pt]
\noindent\textbf{Stochastic Segmentation Networks.} \, Stochastic Segmentation Networks (SSNs)~\citep{monteiro2020stochastic} model joint distributions over label maps to generate spatially coherent segmentations. Pixel-wise dependencies are modelled by placing a low-rank multivariate Gaussian distribution over the logit space, and marginalising\footnote{A Monte Carlo estimator of this (intractable) integral is typically used.} the logits $\boldsymbol{\eta} \in \mathbb{R}^{kd}$ to compute likelihoods:
\begin{align}
    \label{eq: ssn_like}
    p(\mathbf{y} \mid \mathbf{x}) &= \int p(\mathbf{y} \mid \boldsymbol{\eta}) p(\boldsymbol{\eta} \mid \mathbf{x}) \ \mathrm{d}\boldsymbol{\eta}, \nonumber
    \\ p(\boldsymbol{\eta} \mid \mathbf{x}) &= \mathcal{N}(\boldsymbol{\eta};\boldsymbol{\mu}(\mathbf{x}), \boldsymbol{\Sigma}(\mathbf{x})), \nonumber
    \\[2pt] p(\mathbf{y} \mid \boldsymbol{\eta}) &= \mathrm{Categorical}(\mathbf{y};\mathrm{softmax}_k(\boldsymbol{\eta})),
\end{align}
\newpage
\noindent where $\boldsymbol{\mu}(\mathbf{x}) \in \mathbb{R}^{kd}$ and $\boldsymbol{\Sigma}(\mathbf{x}) \in \mathbb{R}^{kd \times kd}$ are predicted by a neural network given an input $\mathbf{x}$. Dependencies in logit space manifest in pixel space through the conditional dependence of $\mathbf{y}$ on $\boldsymbol{\eta}$. Note that $\boldsymbol{\Sigma}(\mathbf{x})$ is not only spatial but also class-wise. Due to the large number of pixels involved, estimating $\boldsymbol{\Sigma}(\mathbf{x})$ is typically computationally infeasible, thus,~\citet{monteiro2020stochastic} use a low-rank approximation of the form:
\begin{align}
    \label{eq: lr_cov}
    \boldsymbol{\Sigma}(\mathbf{x}) = \mathbf{D}(\mathbf{x}) + \mathbf{P}(\mathbf{x})\mathbf{P}(\mathbf{x})^\top,
\end{align}
where $\mathbf{D}(\mathbf{x}) \in \mathbb{R}_{+}^{kd \times kd}$ denotes the diagonal matrix containing pixel-wise (including class-wise) independent variances, whereas the covariance factor matrix specifying $r$ as the rank of the approximation is denoted by $\mathbf{P}(\mathbf{x}) \in \mathbb{R}^{kd \times r}$.

\paragraph{Normalising Flows.} Normalising Flow (NF)~\citep{tabak2013family,dinh2014nice,rezende2015variational,papamakarios2021normalizing} models construct a complex probability distribution $p_X$ of a target variable $X$ by applying a parameterised transformation $\phi: \mathcal{U} \to \mathcal{X}$ to a simple base distribution $p_U$. If the transformation $\mathbf{x} = \phi(\mathbf{u})$ is both invertible and differentiable (i.e. \textit{diffeomorphic}), the density of the target variable is readily given by the change-of-variables formula:
\begin{align}
    \label{eq: change_of_var}
    && p_X(\mathbf{x}) = p_U(\mathbf{u})\left|\det \mathbf{J}_{\phi}(\mathbf{u})\right|^{-1}, && \mathbf{u} = \phi^{-1}(\mathbf{x}),  &&
\end{align}
where $(\mathbf{J}_{\phi}(\mathbf{u}))_{ij} = \partial\phi_i / \partial u_j$ is the Jacobian matrix. The design space of transformations $\phi$ is often restricted to cases where the Jacobian determinant is efficient to compute. Autoregressive flows~\citep{kingma2016improved,papamakarios2017masked} use a chain of invertible transformations $\phi_1 \circ \phi_2 \circ \cdots \circ \phi_T$, each given by an autoregressive model. Autoregressive flows have been used to increase the flexibility of the approximate posterior in VAEs~\citep{kingma2016improved,vahdat2020nvae}.

\paragraph{Continuous Normalising Flows.} 
Continuous Normalising Flows (CNFs)~\citep{chen2018neural} define a time-dependent (for time $t \in [0,1]$) continuous flow mapping $\phi_t(\mathbf{x})$ from a simple base density $\mathbf{x}_0 \sim p_0$ to a desired data distribution $\mathbf{x}_1 \sim p_{\text{data}}$ governed by an ordinary differential equation (ODE):
\begin{align}
    \label{eq: ode}
    && \frac{\mathrm{d}\phi_t(\mathbf{x})}{\mathrm{d}t} = v_t(\phi_t(\mathbf{x});\theta),  && \mathbf{\phi}_0(\mathbf{x}) = \mathbf{x}_0, &&
\end{align}
where $v_t(\phi_t(\mathbf{x});\theta)$ is a vector field parameterised by a deep neural network with parameters $\theta$. A vector field $v_t$ is said to generate a \textit{probability density path} $p_t$ that transports $p_0$ to $p_1 \approx p_{\text{data}}$ if its flow $\phi_t$ satisfies the continuous-time analogue of the change-of-variables formula in Equation~\ref{eq: change_of_var}. To sample from the model, noise is mapped to data by solving the following differential equation using an ODE solver:
\begin{align}
    \label{eq: ode2}
    \mathbf{\phi}_1(\mathbf{x}) = \mathbf{x}_1 = \mathbf{x}_0 + \int_0^1 v_t(\phi_t(\mathbf{x});\theta) \ \mathrm{d}t.
\end{align}
Flow Matching (FM)~\citep{lipman2023flow,liu2023flow,albergo2023building} provides a simulation-free way of training CNFs by regressing a velocity field $u_t$, inducing a desired probability path $p_t$. However, both $p_t$ and $u_t$ are generally unknown; there exist many $p_t$'s which generate $p_{\text{data}}$, and we do not know the $u_t$ that generates $p_t$.
\citet{lipman2023flow} showed that a chosen $p_t$ and corresponding $u_t$ can be constructed by marginalising \textit{conditional} probability paths and vector fields over $p_{\text{data}}$. A conditional probability path $p_t(\mathbf{x} \mid \mathbf{x}_1)$ is a time-dependent distribution that satisfies the following marginal constraints at the endpoints:
\begin{align}
    &&p_{0}(\mathbf{x} \mid \mathbf{x}_1) = p_0(\mathbf{x}), && p_1(\mathbf{x} \mid \mathbf{x}_1) = \delta(\mathbf{x} - \mathbf{x}_1). &&
\end{align}
The marginal probability path and vector field are given by:
\begin{align}
    p_t(\mathbf{x}) &= \mathbb{E}_{\mathbf{x}_1 \sim p_{\text{data}}} \left[p_t(\mathbf{x} \mid \mathbf{x}_1)\right] \nonumber
    \\ u_t(\mathbf{x}) &= 
    \mathbb{E}_{\mathbf{x}_1 \sim p_{\text{data}}} \left[u_t(\mathbf{x} \mid \mathbf{x}_1)\frac{p_t(\mathbf{x} \mid \mathbf{x}_1)}{p_t(\mathbf{x})} \right],
\end{align}
where the \textit{conditional} vector field $u_t(\mathbf{x} \mid \mathbf{x}_1)$ is defined by the time derivative $\mathrm{d}\phi_t/\mathrm{d}t$ of the chosen flow map $\phi_t$, which transports samples from $p_0$ to $p_t(\mathbf{x} \mid \mathbf{x}_1)$. A common choice is to set the flow to $\phi_t(\mathbf{x}_0 \mid \mathbf{x}_1) = \sigma_t(\mathbf{x}_1)\mathbf{x}_0 + \mu_t(\mathbf{x}_1)$, where $p_t(\mathbf{x} \mid \mathbf{x}_1) = \mathcal{N}(\mathbf{x}; \mu_t(\mathbf{x}_1), \sigma_t^2(\mathbf{x}_1)I)$ is Gaussian.

Crucially, \citet{lipman2023flow} showed that if $u_t(\mathbf{x} \mid \mathbf{x}_1)$ generates $p_t(\mathbf{x} \mid \mathbf{x}_1)$ then $u_t(\mathbf{x})$ generates $p_t(\mathbf{x})$ and the following simple regression objective can be used to train a CNF that generates the marginal probability path $p_t(\mathbf{x})$:
\begin{align}
    \mathbb{E}_{t, p_{\text{data}}(\mathbf{x}_1), p_t(\mathbf{x} \mid \mathbf{x}_1)}\left[\| u_t(\mathbf{x} \mid \mathbf{x}_1) - v_t(\mathbf{x}; \theta ) \|^2\right],
\end{align}
with $t \sim \mathcal{U}(0,1)$, thus approximating the unknown data distribution $p_1 \approx p_{\text{data}}$ at time $t=1$, as intended. 
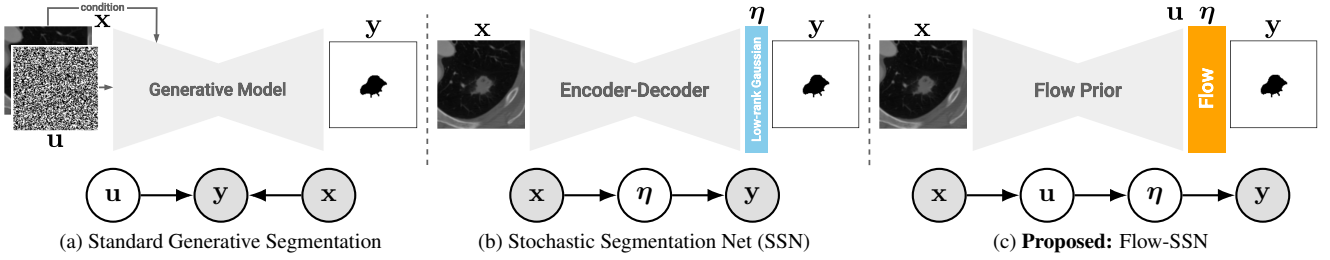
\begin{figure*}[!ht]
    \includegraphics[width=\textwidth]{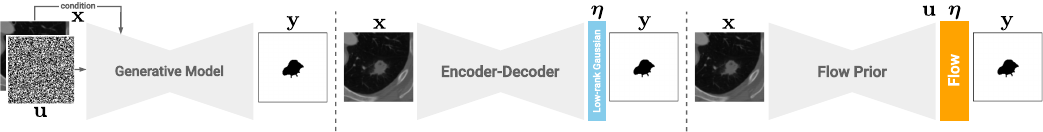}
    \hfill
    \begin{subfigure}{.33\textwidth}
        \centering        
        \begin{tikzpicture}[thick,scale=1, every node/.style={scale=1}]
            \node[latent] (u) {$\mathbf{u}$};
            \node[obs, right=20pt of u] (y) {$\mathbf{y}$};
            \node[obs, right=20pt of y] (x) {$\mathbf{x}$};
    
            \edge[-{Latex[scale=1.0]}]{u}{y}
            \edge[-{Latex[scale=1.0]}]{x}{y} 
        \end{tikzpicture}
        \caption{Standard Generative Segmentation}
    \end{subfigure}
    \begin{subfigure}{.305\textwidth}
        \centering
        \begin{tikzpicture}[thick,scale=1, every node/.style={scale=1}]
            \node[obs] (x) {$\mathbf{x}$};
            \node[latent, right=20pt of u] (eta) {$\boldsymbol{\eta}$};
            \node[obs, right=20pt of eta] (y) {$\mathbf{y}$};
    
            \edge[-{Latex[scale=1.0]}]{x}{eta}
            \edge[-{Latex[scale=1.0]}]{eta}{y} 
        \end{tikzpicture}
        \caption{Stochastic Segmentation Net (SSN)}
    \end{subfigure}
    \hfill
    \begin{subfigure}{.33\textwidth}
        \centering
        \begin{tikzpicture}[thick,scale=1, every node/.style={scale=1}]
            \node[obs] (x) {$\mathbf{x}$};
            \node[latent, right=20pt of x] (u) {$\mathbf{u}$};
            \node[latent, right=20pt of u] (eta) {$\boldsymbol{\eta}$};
            \node[obs, right=20pt of eta] (y) {$\mathbf{y}$};
        
            \edge[-{Latex[scale=1.0]}]{x}{u}
            \edge[-{Latex[scale=1.0]}]{u}{eta}
            \edge[-{Latex[scale=1.0]}]{eta}{y} 
        \end{tikzpicture}
        \caption{\textbf{Proposed:} Flow-SSN}
    \end{subfigure}
    \caption{\textbf{Graphical models for stochastic segmentation}. \textbf{(a)} The typical setup, a generative model of label maps $\mathbf{y}$ conditioned on the image $\mathbf{x}$ (e.g. diffusion-based segmentation).
    \textbf{(b)} Stochastic segmentation network~\cite{monteiro2020stochastic}, a Markov chain with a low-rank Gaussian parameterisation of the logits $\boldsymbol{\eta}$, where $\mathbf{y} = \mathrm{softmax}(\boldsymbol{\eta})$. \textbf{(c)} A Flow-SSN (discrete or continuous-time), which consists of: (i) a parameterised conditional base distribution
    $p_{U|X}(\mathbf{u} \mid \mathbf{x}; \lambda)$ serving as an expressive flow \textit{prior}; (ii) a lightweight flow $\phi : \mathcal{U} \to \mathcal{Y}$ to model pixel-wise dependencies.
    }
    \label{fig: pgms}
\end{figure*}
\section{Theoretical Analysis: Effective Rank} 
\label{sec: motivation}
As outlined in Section~\ref{sec: introduction}, the low-rank assumption in SSNs is restrictive.
Even if scaling the number of covariance factors (e.g. $r \gg 10$) were practical, it would still impose a Gaussian assumption on the pixel-space distribution.
We now proceed with a theoretical analysis of the rank assumption in SSNs, to provide a more nuanced argument in favour of our flow-based approach. In short, we prove that the expected likelihood SSNs use results in a rank \textit{increase} relative to the initially assumed rank, but the final effective rank only grows sublinearly w.r.t. the initial rank, limiting expressivity\footnote{Proofs for all formal results are provided in Appendices 
A and B.
}.

\begin{lemma}[Rank Increase]
\label{prop: rank_increase}
Let the logits $\boldsymbol{\eta}$ be low-rank Gaussian distributed $\boldsymbol{\eta} \mid \mathbf{x} \sim \mathcal{N}(\boldsymbol{\mu}(\mathbf{x}), \boldsymbol{\Sigma}(\mathbf{x}))$, where the covariance $\boldsymbol{\Sigma}(\mathbf{x}) \in \mathbb{R}^{kd \times kd}$ has rank $\mathrm{rank}(\boldsymbol{\Sigma}(\mathbf{x})) = r$. Given that $\mathbf{y} =\mathrm{softmax}_k(\boldsymbol{\eta})$, the following holds:
\begin{align}
    \mathrm{rank}(
    \mathrm{Cov}(\mathbf{y})) 
    > r \iff r < d(k-1).
\end{align}
\end{lemma}
{\noindent 
This result reveals that the low-rank approximation is not as restrictive as anticipated, as the non-linear pushforward by the softmax induces a rank \textit{increase}. However, we now prove that the \textit{effective} rank remains low, limiting expressivity.}

\begin{definition}[Effective Rank~\cite{roy2007effective}] 
\label{def: effective_rank}
The effective rank $\mathrm{erank}(A) \in \mathbb{R}$ of a matrix $A \in \mathbb{R}^{d \times d}$ is given by:
\begin{align}
    &&\mathrm{erank}(A) = e^{H(p)}, && H(p) = -\sum_{i=1}^d p_i \log p_i, &&
\end{align}
where $H(p)$ is the Shannon entropy of the singular value distribution: $p_i = \sigma_i / \sum_{j=1}^d \sigma_j$, for $i \in \{1,\dots,d\}$, with $\sigma_1 \geq \sigma_2 \geq \dots \geq \sigma_d \geq 0$ denoting the singular values of $A$.
\end{definition}
\begin{remark}
    Intuitively, the effective rank represents the average number of significant dimensions in a matrix's range.
\end{remark}

\begin{theorem}[Sublinear Growth of the Effective Rank]
\label{prop: sublin_main}
Given a low-rank Gaussian covariance matrix $\boldsymbol{\Sigma}(\mathbf{x}) \in \mathbb{R}^{kd \times kd}$ with initial rank $\mathrm{rank}(\boldsymbol{\Sigma}(\mathbf{x})) = r < d(k-1)$. The increase in the effective rank $\mathrm{erank}(\mathrm{Cov}(\mathbf{y}))$, in the sense of Lemma~\ref{prop: rank_increase}, grows sublinearly w.r.t. $r$.
\end{theorem}
\noindent Theorem~\ref{prop: sublin_main} shows that despite the induced rank increase as per Lemma~\ref{prop: rank_increase}, the final effective rank $\mathrm{erank}(\mathrm{Cov}(\mathbf{y}))$ only grows sublinearly w.r.t the initial rank $r$. Thus, the low-rank assumption is still highly restrictive for high-dimensional, pixel-wise covariances. This result may be of independent interest to other low-rank approximation techniques~\cite{yang2018breaking,hu2022lora}.

\section{Flow Stochastic Segmentation Networks}
\label{sec: flow ssn}
Motivated by the theoretical analysis in Section~\ref{sec: motivation}, we present Flow Stochastic Segmentation Networks (Flow-SSNs). Unlike standard SSNs, Flow-SSNs can estimate arbitrarily high-rank pixel-wise covariances without needing to assume the rank apriori or store distributional parameters. We begin our exposition by: (\S\ref{subsec: discrete_flows}) designing \textit{discrete-time} flows for learning pixel-wise covariances from an autoregressive perspective, then (\S\ref{subsec: Continuous Flows for Stochastic Segmentation}) develop a generalisation of the approach to modern \textit{continuous-time} flows, which admit free-form Jacobians. Flow-SSNs are also significantly more efficient to sample from than other diffusion-based segmentation models, as the majority of the model capacity is allocated to learning a flow's prior (normally fixed), while the flow network is lightweight, thereby reducing sampling cost substantially.
\newpage
\subsection{Discrete-Time Autoregressive Flow-SSNs} 
\label{subsec: discrete_flows}
The motivation for using flows to model pixel-wise covariances in stochastic segmentation is simple. Since pixel-wise Gaussian covariances only represent linear dependencies between components, a \textit{linear} autoregressive flow model is sufficient to transform a Gaussian distributed random variable with diagonal covariance into one with \textit{full} covariance.
\begin{proposition}
[Full Covariance Flow Transformation]
\label{prop: full_cov}
Let $\mathbf{u} = (u_1, u_2, \dots, u_d)^{\top}$ be a Gaussian distributed random vector with diagonal covariance $\mathbf{u} \sim \mathcal{N}(\boldsymbol{\mu}, \mathrm{diag}(\boldsymbol{\sigma}^2))$. A linear autoregressive model is sufficient to transform $\mathbf{u}$ into a new variable $\boldsymbol{\eta} \in \mathbb{R}^d$ with full covariance $\boldsymbol{\Sigma} \in \mathbb{R}^{d \times d}$.
\end{proposition}
\noindent This result underpins our key model proposal, which consists of the following two components: (i) learn an expressive, but pixel-wise independent, flow base distribution conditional on $\mathbf{x}$ to act as an `initial guess' logit distribution; (ii) use a lightweight autoregressive flow to model logit-space dependencies, thereby inducing \textit{full} covariance structure as per Proposition~\ref{prop: full_cov}, and refining the initial guess. Concretely, a Flow-SSN consists of the following two components: 

\begin{enumerate}[{(i)}]
    \item A conditional base distribution $p_{U|X}$, parameterised by a neural encoder-decoder $f_\lambda : \mathbb{R}^{c\times d} \to \mathbb{R}^{kd} \times \mathbb{R}^{kd}$:
    \begin{align}
        p_{U|X}(\mathbf{u} \mid \mathbf{x}; \lambda) = \mathcal{N}\left(\mathbf{u};\boldsymbol{\mu}(\mathbf{x}), \mathrm{diag}(\boldsymbol{\sigma}^2(\mathbf{x}))\right);
    \end{align}
    \item A lightweight autoregressive flow 
    $\phi : \mathbb{R}^{kd} \to \mathbb{R}^{kd}$:
    \begin{align}
        &&\forall i \ : \ \eta_i = \phi_i(\mathbf{u}_{\leq i};\theta), 
        && \mathbf{u}| \mathbf{x} \sim p_{U|X}.&&
\end{align}
\end{enumerate}

\noindent The likelihood $p(\mathbf{y} \mid \mathbf{x})$ of a Flow-SSN can be obtained by marginalising the logit variables $\boldsymbol{\eta}$ similar to an SSN:
\begin{align}
    p(\mathbf{y} \mid \mathbf{x}) &= \int p(\mathbf{y} \mid \boldsymbol{\eta}) p(\boldsymbol{\eta} \mid \mathbf{x};\lambda,\theta) \ \mathrm{d}\boldsymbol{\eta}, \label{eq: flow_ssn} \nonumber
    \\ p(\boldsymbol{\eta} \mid \mathbf{x};\lambda, \theta) &= p_{U|X}(\mathbf{u} \mid \mathbf{x}; \lambda) \left|\det \mathbf{J}_{\phi}(\mathbf{u})\right|^{-1}, \nonumber
    \\[4pt] p(\mathbf{y} \mid \boldsymbol{\eta}) &= \mathrm{Categorical}(\mathbf{y};\mathrm{softmax}_k(\boldsymbol{\eta})),
\end{align}
but $p(\boldsymbol{\eta} \mid \mathbf{x};\lambda, \theta)$ is now more expressive than the low-rank Gaussian in SSNs, as it can model \textit{full} covariance structure.

\subsubsection{Designing a Flow \& Objective}
\begin{figure}[!t]
    \centering
    \begin{subfigure}{.49\columnwidth}
        \centering
        \begin{tikzpicture}[thick,scale=1, every node/.style={scale=1}]
            \node[latent] (u1) {$\mathbf{u}_1$};
            \node[latent, right=15pt of u1] (x1) {$\boldsymbol{\eta}_1$};
            \node[latent, below=15pt of u1] (u2) {$\mathbf{u}_2$};
            \node[latent, right=15pt of u2] (x2) {$\boldsymbol{\eta}_2$};
            \node[latent, below=30pt of u2] (uT) {$\mathbf{u}_d$};
            \node[latent, right=15pt of uT] (xT) {$\boldsymbol{\eta}_d$};
        
            \edge[-{Latex[scale=1.0]}]{u1}{x1}
            \edge[-{Latex[scale=1.0]}]{u2}{x2}
            \edge[-{Latex[scale=1.0]}]{uT}{xT}
        
            \edge[-{Latex[scale=1.0]}]{u1}{u2}
            \draw [-{Latex[scale=1.0]}] (u1) to [out=-45,in=45] (uT);
            \draw [-{Latex[scale=1.0]}] (u2) to [out=-45,in=45] (uT);
            
            \draw node[draw=none, scale=0.75, below=5pt of u2] (u3) {\hspace{0.5pt}\rotatebox{90}{$\mathbf{\cdots}$}};
        
            \edge[-]{u2}{u3}
            \edge[-{Latex[scale=1.0]}]{u3}{uT} 
            
            \node[obs, left=15pt of u2] (x) {$\mathbf{x}$};
            \edge[-{Latex[scale=1.0]}]{x}{u1}
            \edge[-{Latex[scale=1.0]}]{x}{u2}
            \edge[-{Latex[scale=1.0]}]{x}{uT}
        \end{tikzpicture}
    \end{subfigure}
    \begin{subfigure}{.49\columnwidth}
        \centering
        \begin{tikzpicture}[thick,scale=1, every node/.style={scale=1}]
            \node[latent] (u1) {$\mathbf{u}_1$};
            \node[latent, right=15pt of u1] (x1) {$\boldsymbol{\eta}_1$};
            \node[latent, below=15pt of u1] (u2) {$\mathbf{u}_2$};
            \node[latent, right=15pt of u2] (x2) {$\boldsymbol{\eta}_2$};
            \node[latent, below=30pt of u2] (uT) {$\mathbf{u}_d$};
            \node[latent, right=15pt of uT] (xT) {$\boldsymbol{\eta}_d$};
        
            \edge[-{Latex[scale=1.0]}]{u1}{x1}
            \edge[-{Latex[scale=1.0]}]{u2}{x2}
            \edge[-{Latex[scale=1.0]}]{uT}{xT}
        
            \edge[-{Latex[scale=1.0]}]{x1}{x2}
            \draw [-{Latex[scale=1.0]}] (x1) to [out=-45,in=45] (xT);
            \draw [-{Latex[scale=1.0]}] (x2) to [out=-45,in=45] (xT);
        
            \draw node[draw=none, scale=0.75, below=5pt of x2] (x3) {\hspace{0.5pt}\rotatebox{90}{$\mathbf{\cdots}$}};
            \edge[-]{x2}{x3}
            \edge[-{Latex[scale=1.0]}]{x3}{xT}  
            
            \node[obs, left=15pt of u2] (x) {$\mathbf{x}$};
            \edge[-{Latex[scale=1.0]}]{x}{u1}
            \edge[-{Latex[scale=1.0]}]{x}{u2}
            \edge[-{Latex[scale=1.0]}]{x}{uT}
        \end{tikzpicture}
    \end{subfigure}
    \caption{\textbf{Graphical models of autoregressive Flow-SSNs}. \textit{Left}: inverse autoregressive; \textit{Right}: masked autoregressive. The target variable $\mathbf{y}$ is omitted here for simplicity, noting that the logits $ \boldsymbol{\eta} \to \mathbf{y}$ in all cases, e.g. via a deterministic transform $\mathbf{y} = \mathrm{softmax}_k(\boldsymbol{\eta})$.
    }
    \label{fig: gm_flowssn}
\end{figure}
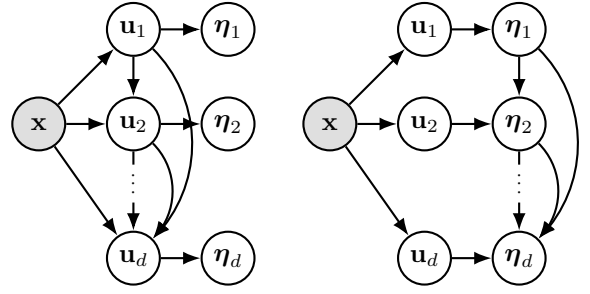

In this section, we explore the design space of autoregressive Flow-SSNs. Affine autoregressive flows are of the form $\eta_i = \phi_{\mu_i}(\boldsymbol{\eta}_{<i}) + \phi_{\sigma_i}(\boldsymbol{\eta}_{<i}) u_i$, and 
have a tractable, lower triangular Jacobian: $\partial u_i/\partial \eta_j = 0, \forall j>i$.
Using a Masked Autoregressive Flow (MAF)~\citep{papamakarios2017masked} for modelling $p(\boldsymbol{\eta} \mid \mathbf{x})$ is not ideal, as it requires slow, pixel-wise sequential sampling. Conversely, Inverse Autoregressive Flows (IAFs)~\citep{kingma2016improved} are fast to sample from, but require pixel-wise sequential scoring\footnote{As per Figure~\ref{fig: gm_flowssn}, the inverse transform of an affine autoregressive flow $\mathbf{u} = (\boldsymbol{\eta} - \boldsymbol{\mu}(\boldsymbol{\eta})) / \boldsymbol{\sigma}(\boldsymbol{\eta})$ is parallelisable since $u_i \perp\!\!\!\perp u_{j \neq i}\mid \{\eta_i, \mu_i, \sigma_i\}$.}. Crucially, IAFs can still score their \textit{own} samples efficiently, as intermediate outputs can be cached. 

This important fact opens up multiple design options for building discrete-time autoregressive Flow-SSNs, some of which we outline next and detail further in Appendix
B.
\\[4pt]
\noindent\textbf{Inverse Autoregressive Flow-SSN.} 
A simple approach is to choose an IAF, and use a Monte Carlo estimator of the categorical likelihood in Eq.~\eqref{eq: flow_ssn} analogous to standard SSNs:
\begin{align}
    \max_{\lambda,\theta} \mathbb{E}_{\mathbf{u} \sim p_{U|X}(\mathbf{u}|\mathbf{x};\lambda)}\left[\log p(\mathbf{y} \mid \boldsymbol{\eta}=\phi(\mathbf{u};\theta))\right].
    \label{eq:discrete_xlike}
\end{align}
\textbf{Dual Flow-SSN.} A \textit{dual} Flow-SSN comprised of an IAF $p^{\text{IAF}}(\boldsymbol{\eta} \mid \mathbf{x}; \lambda, \theta)$ and an MAF $p^{\text{MAF}}(\boldsymbol{\eta} \mid \mathbf{x}; \hat{\lambda}, \hat{\theta})$ can be trained concurrently by maximising: $\log p(\mathbf{y} \mid \mathbf{x}) \geq$
\begin{align}
    \mathbb{E}_{\boldsymbol{\eta} \sim p^{\text{IAF}}(\boldsymbol{\eta} \mid \mathbf{x};\lambda, \theta)}\left[\log p(\mathbf{y} \mid \boldsymbol{\eta})\right] \nonumber
    - D_{\mathrm{KL}}\left(p^{\text{IAF}} \parallel p^{\text{MAF}}\right).
\end{align}
If we choose $p^{\text{MAF}}$ as improper uniform: $\forall \boldsymbol{\eta}, p(\boldsymbol{\eta}) = \mathrm{const}$, then we can avoid training the MAF 
by optimising:
\begin{align}    \mathbb{E}_{\boldsymbol{\eta} \sim p^{\text{IAF}}(\boldsymbol{\eta} \mid \mathbf{x};\lambda, \theta)}\left[\log p(\mathbf{y} \mid \boldsymbol{\eta})\right] + \beta H(p^{\text{IAF}}),
\end{align}
where setting the hyperparameter $\beta > 0$ helps prevent $p^{\text{IAF}}$ from collapsing to a deterministic model. 

\subsection{Continuous-Time Flow-SSNs}
\label{subsec: Continuous Flows for Stochastic Segmentation}
Relaxing the autoregressive structure of discrete-time Flow-SSNs induces a more expressive \textit{free-form} Jacobian, but it complicates the computation of its determinant, as it is no longer simply the product of its diagonal elements. This calls for adapting continuous-time flows~\citep{chen2018neural,lipman2023flow,liu2023flow,albergo2023building} to build Flow-SSNs, as they can be trained efficiently via FM, admit free-form Jacobians, and can therefore model arbitrary pixel-wise covariances in stochastic segmentation tasks.
\\[4pt]
\noindent\textbf{Interpolation Path.} Continuous-time Flow-SSNs share the same design principle as their discrete-time counterparts, comprising: (i) an expressive conditional base distribution; (ii) a lightweight flow to model pixel-wise dependencies. However, the flow $\phi_t$ is now a continuous-time mapping from the conditional base distribution $\mathbf{u}|\mathbf{x} \sim p_{t=0} = p_{U|X}$ to the label data distribution $\mathbf{y} \sim p_{t=1} = p_{\text{data}}$, and it can be defined as a simple deterministic interpolation path:
\begin{align}
    \mathbf{y}_t = (1-t)\mathbf{u} + t\mathbf{y} \implies 
    \mathrm{d}\mathbf{y}_t = \left(\mathbf{y} - \mathbf{u}\right)\mathrm{d}t.
\end{align}
Importantly, the so-called \textit{Markovian projection} $\mathcal{M}(\cdot)$~\cite{gyongy1986mimicking} of this type of ODE has been previously shown~\cite{peluchetti2022nondenoising,shi2024diffusion} to preserve the marginals $p_t$ for all time points $t \in [0,1]$:
\begin{align}
    \mathrm{d}\mathbf{y}^\star_t = 
    \mathcal{M}\left(\mathbf{y} - \mathbf{u}\right)\mathrm{d}t = 
    \left(\mathbb{E}[\mathbf{y}\mid\mathbf{y}_t] - \mathbf{u}\right)\mathrm{d}t.
\end{align}
Therefore, in practice, we require a model $p(\cdot;\theta)$ to approximate the conditional expectation $\mathbb{E}[\mathbf{y}\mid\mathbf{y}_t]$, which we propose to parameterise using a \textit{much smaller} neural network compared to the one used for learning the base distribution $p_{U|X}$ (i.e. prior).
\noindent This makes sampling (ODE solving) \textit{significantly} cheaper compared to typical generative segmentation models, where \textit{all} the model parameters are dedicated to learning the score/velocity field~\cite{amit2021segdiff,rahman2023ambiguous,zbinden2023stochastic,wu2024medsegdiff}. Further, one could then leverage a large foundation model as the prior.
\\[4pt]
\noindent\textbf{Categorical Likelihood.} Since $\mathbf{y}$ is a one-hot map, we have:
\begin{align}
    \mathbb{E}[\mathbf{y} \mid \mathbf{y}_t] 
    &= \sum_{\mathbf{y} \in \{0, 1\}^{k \times d}} \mathbf{y} \cdot \mathrm{Categorical}(\mathbf{y};\mathrm{softmax}(\boldsymbol{\eta})) \nonumber
    \\ &\approx \mathrm{softmax}(\boldsymbol{\eta}(\phi_t(\mathbf{u} \mid \mathbf{y});\theta)),
\end{align}
where $\phi_t(\mathbf{u}\mid \mathbf{y}) = \mathbf{y}_t = (1-t)\mathbf{u} + t\mathbf{y}$, and $\boldsymbol{\eta}(\cdot;\theta)$ is a neural network.
Thus, like discrete-time Flow-SSNs (cf. Eq.~\eqref{eq:discrete_xlike}), we train using an expected categorical likelihood:
\begin{align}
    \max_{\lambda, \theta} \mathbb{E}_{\mathbf{u} \sim p_{U|X}(\mathbf{u}|\mathbf{x};\lambda)} \left[\int_0^1\log p(\mathbf{y} \mid \mathbf{y}_t;\theta) \ \mathrm{d}t\right],
\end{align}
where $(\mathbf{x}, \mathbf{y}) \sim p_{\text{data}}(\mathbf{x}, \mathbf{y})$, and we now have $t \sim \mathcal{U}(0,1)$. This objective differs from the standard FM objective and can be seen as a special case of variational FM~\cite{eijkelboom2024variational}. However, the motivation and derivations presented here are distinct, as they are a natural consequence of infinite-depth Flow-SSNs. 
\\[5pt]
\noindent\textbf{Priors.} Alternative priors for continuous-time Flow-SSNs can be specified by, e.g., a pushforward of the base distribution $f_\#p_{U|X}$, then defining the interpolant $\mathrm{d}\mathbf{y}_t = (\mathbf{y} - \boldsymbol{\eta}) \mathrm{d}t$. For instance, if we choose $\boldsymbol{\eta} = f(\mathbf{u})$ as the log-softmax function, then $f_\#p_{U|X}$ is log-logistic normal. Another promising avenue is to consider mixture distributions (e.g. Gaussian) and/or leverage large foundation models as flexible priors.
\begin{figure}[t]
    \centering
    \begin{subfigure}{.38\columnwidth}
        \centering\includegraphics[width=\columnwidth]{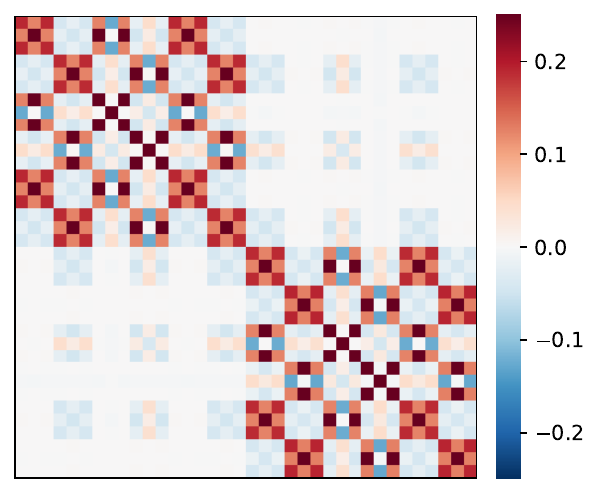}
    \end{subfigure}
    \hfill
    \begin{subfigure}{.61\columnwidth}
        \centering
        \includegraphics[
        trim={0 0 0 0},clip, width=\columnwidth
        ]{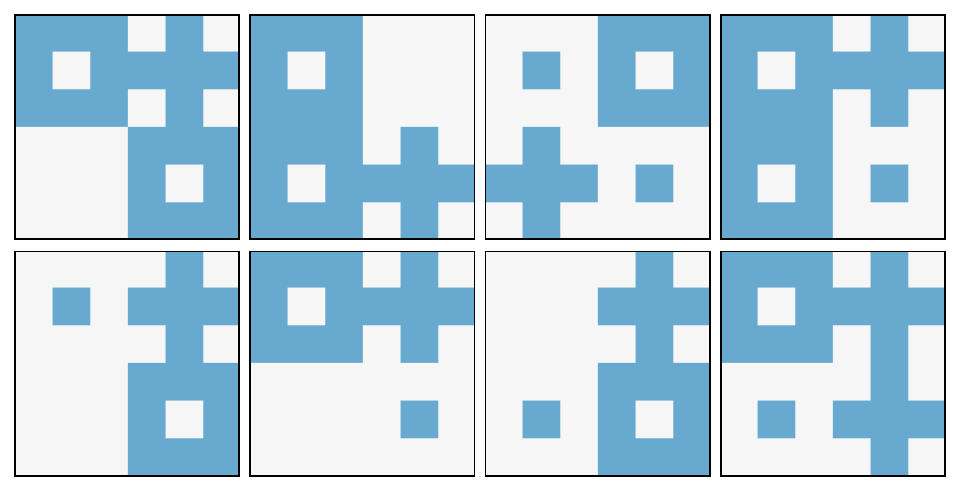}
    \end{subfigure}
    \caption{\textbf{The \textit{MarkovShapes} dataset}. (\textit{Left}) Ground truth pixel covariance matrix, rank $r_{\text{true}}{=}12$. (\textit{Right}) Random data samples.}
    \label{fig: markovshapes}
\end{figure}
\section{Experiments}
\label{sec: experiments}
\subsection{Toy Problem: MarkovShapes}
\label{subsec: toy}
In this experiment, we introduce a synthetic dataset, \textit{MarkovShapes}, where we have full control of the rank of the pixel-space covariance. Comparing Flow-SSN with the low-rank SSN model~\cite{monteiro2020stochastic}, \textit{MarkovShapes} reveals how the SSN fails when the assumed rank is underspecified and demonstrates how Flow-SSN overcomes this shortcoming. 

\textit{MarkovShapes} consists of images composed of 3 possible shapes: `\textit{square}' (\includegraphics[width=0.03\columnwidth]{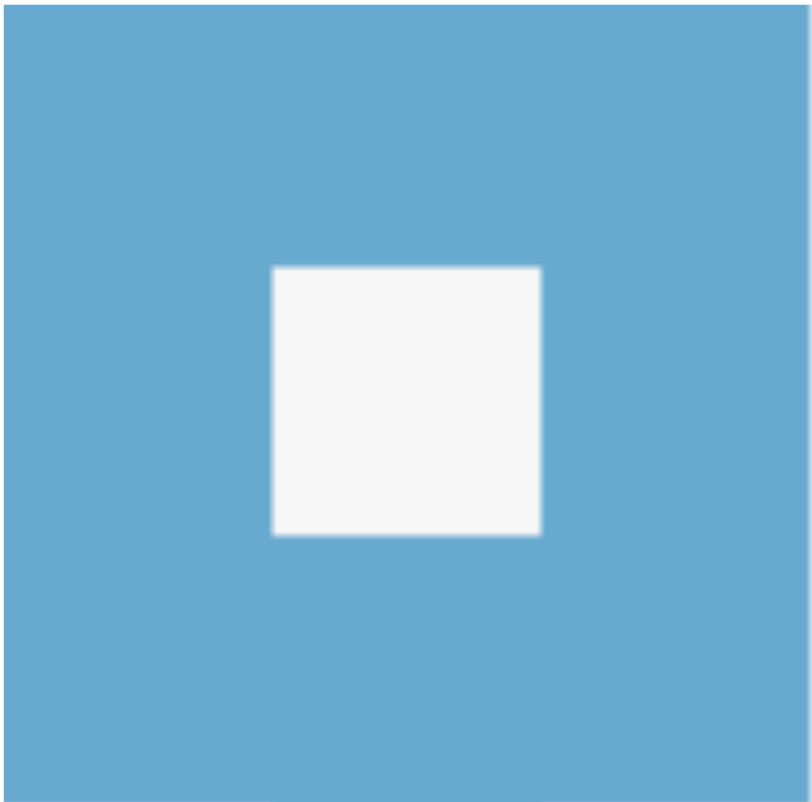}), `\textit{plus}' (\includegraphics[width=0.03\columnwidth]{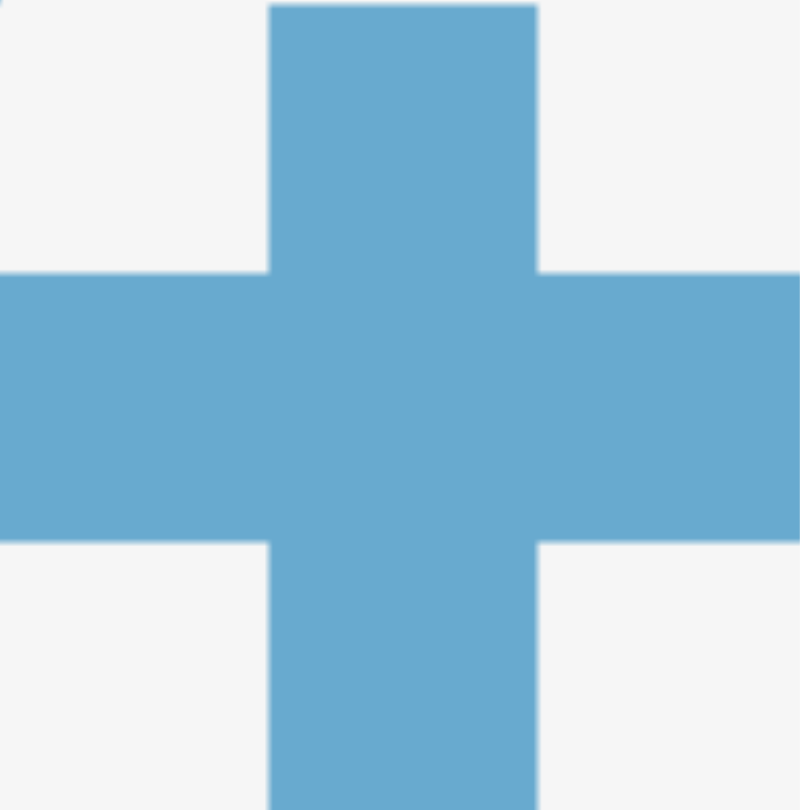}) and `\textit{dot}'(\includegraphics[width=0.03\columnwidth]{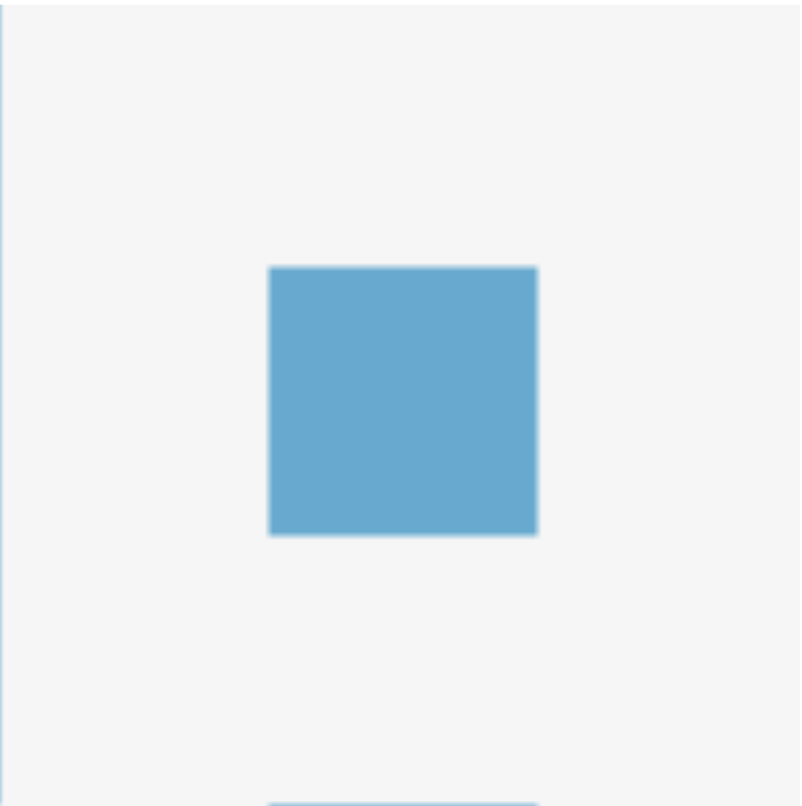}). Each image quadrant is either filled with a shape or left empty ($\varnothing$) at random. The data-generating process is a Markov chain, with states corresponding to shapes and the transitions between quadrants governed by the doubly stochastic matrix $\mathbf{T}$:
\begin{align}
    \setlength{\bigstrutjot}{1ex}
    \mathbf{T} = 
    \begin{blockarray}{ccccc}
         & \varnothing  & \includegraphics[width=0.04\columnwidth]{images/square.png}  & \includegraphics[width=0.04\columnwidth]{images/plus.png} & \includegraphics[width=0.04\columnwidth]{images/dot.png} 
         \\
        \begin{block}{l[>{\medspace}c c c c<{\medspace}]}
            \hspace{1pt} \varnothing & \nicefrac{1}{4} & \nicefrac{1}{4} & \nicefrac{1}{4} & \nicefrac{1}{4}\bigstrut[t] 
            \\
            \includegraphics[width=0.04\columnwidth]{images/square.png} &  \nicefrac{1}{4} & \nicefrac{3}{40} & \nicefrac{27}{80} & \nicefrac{27}{80} 
            \\
            \includegraphics[width=0.04\columnwidth]{images/plus.png} &  \nicefrac{1}{4} & \nicefrac{27}{80} & \nicefrac{3}{40} & \nicefrac{27}{80} 
            \\
            \includegraphics[width=0.04\columnwidth]{images/dot.png} & \nicefrac{1}{4} & \nicefrac{27}{80} & \nicefrac{27}{80} & \nicefrac{3}{40}\bigstrut[b]
            \\[-5pt]
        \end{block}
    \end{blockarray}, 
    \quad
\end{align}
where $\mathbf{T}_{ij} = P(X_{t+1} = j \mid X_{t} = i)$ is the probability of transitioning from shape $i$ to $j$ within a particular image. Figure~\ref{fig: markovshapes} shows the induced pixel covariance matrix, its empirically calculated true rank $r_{\text{true}}$, and random samples.

\begin{figure}[!t]
    \centering
    \includegraphics[width=\columnwidth]{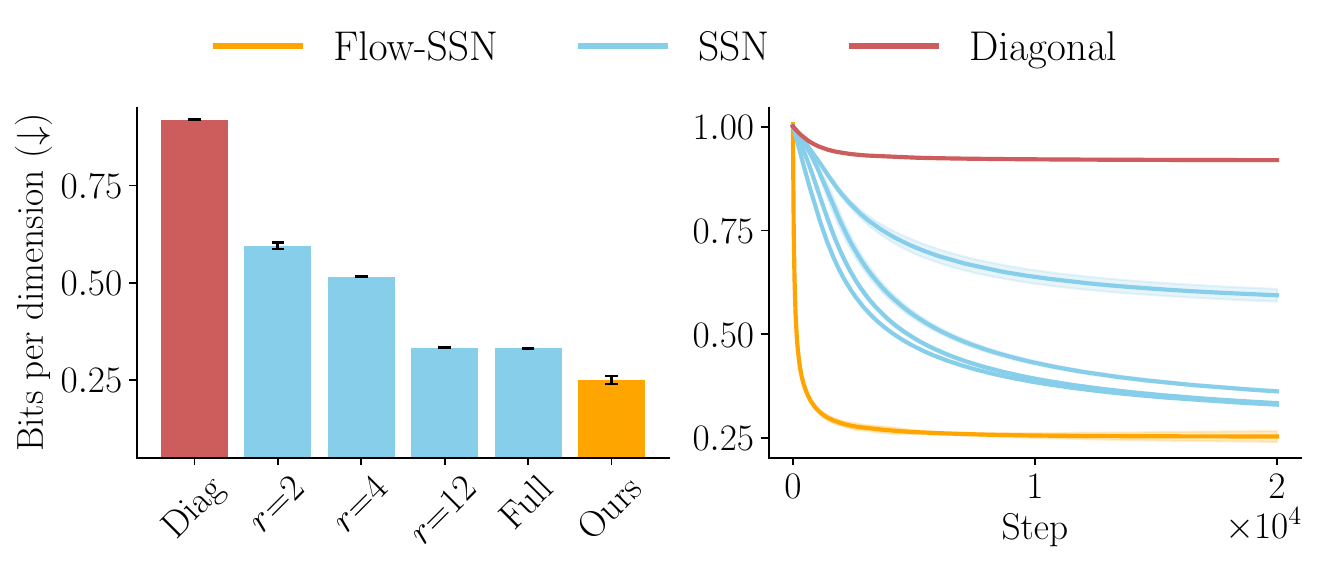}
    \caption{Bits per dimension (BPD) results on \textit{MarkovShapes}.}
    \label{fig:markovshapes-bpd}
\end{figure}

\begin{figure}[!t]
    \centering
    \begin{subfigure}{.38\columnwidth}
        \centering
        \includegraphics[width=\columnwidth]{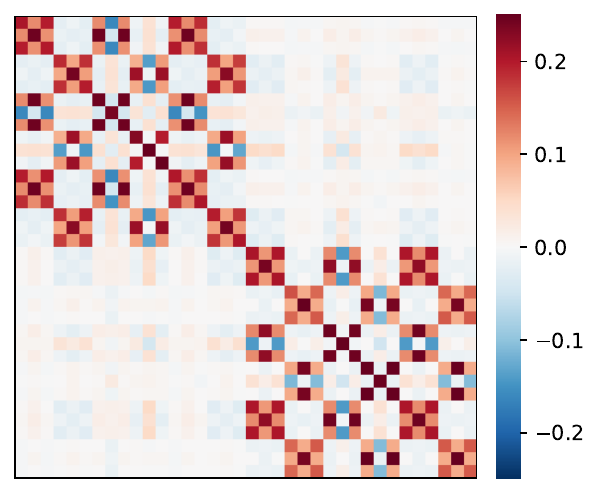}
        \captionsetup{labelfont=bf}
        \caption{Cov (Flow-SSN)}
    \end{subfigure}
    \hfill
    \begin{subfigure}{.61\columnwidth}
        \centering
        \includegraphics[trim={0 0 0 0},clip, width=\columnwidth]{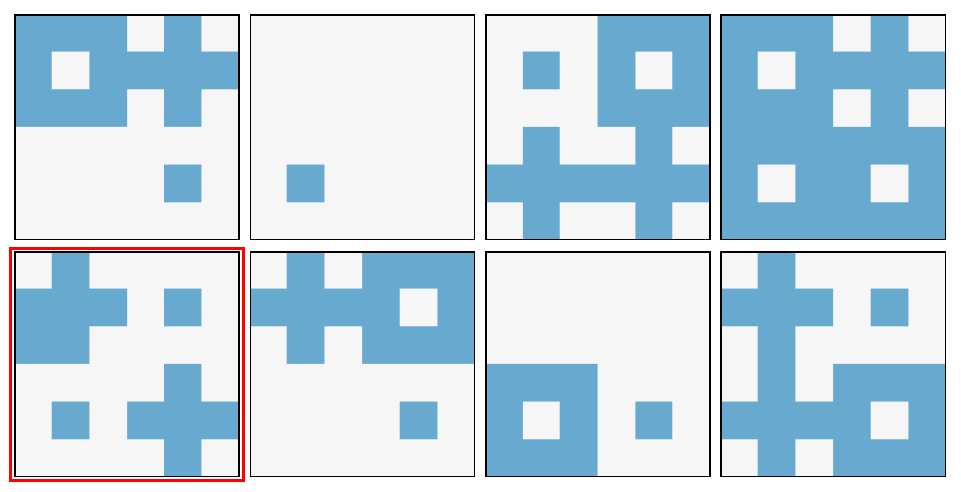}
        \captionsetup{labelfont=bf}
        \caption{Samples (Flow-SSN)}
    \end{subfigure}
    \\[4pt]
    \begin{subfigure}{.38\columnwidth}
        \centering
        \includegraphics[width=\columnwidth]{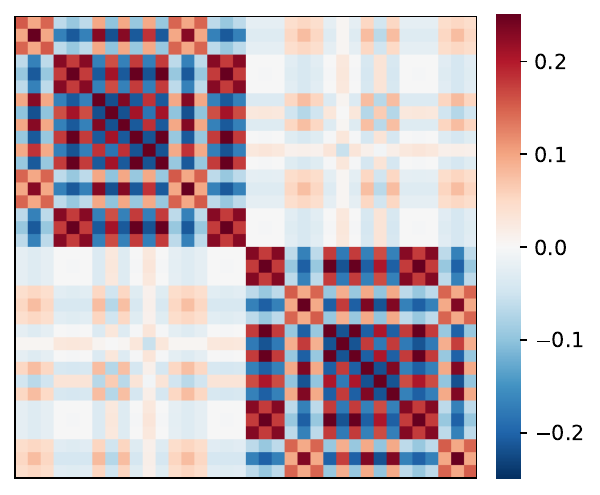}
        \captionsetup{labelfont=bf}
        \caption{Cov (SSN $r{=}2$)}
    \end{subfigure}
    \hfill
    \begin{subfigure}{.61\columnwidth}
        \centering
        \includegraphics[trim={0 0 0 0},clip, width=\columnwidth]{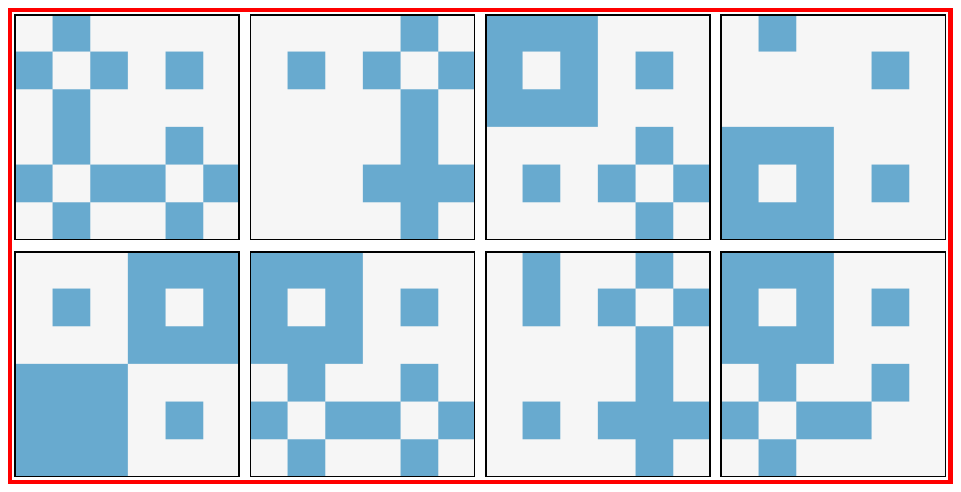}
        \captionsetup{labelfont=bf}
        \caption{Samples (SSN $r{=}2$)}
    \end{subfigure}
    \caption{\textbf{Comparing learnt covariances on \textit{MarkovShapes}}. (\textit{Top}) Flow-SSNs make no rank assumptions and approximate the ground truth covariance well. (\textit{Bottom}) The low-rank approximation causes sampling errors, even under a mild underspecification ratio, $r{=}2$ relative to $r_{\text{true}}{=}12$. Sampling errors (hallucinated shapes not faithful to the true covariance structure) are indicated with red boxes.}
    \label{fig:markovshapes-samples}
\end{figure}

We adapt the SSN from the toy problem of \citet{monteiro2020stochastic}, training four variants from rank 2 to full-rank on \textit{MarkovShapes}. We implement and train a discrete-time autoregressive Flow-SSN with the objective in Eq.~\eqref{eq:discrete_xlike}. We choose a single linear layer with MADE-style masking~\cite{germain2015made} for the autoregressive flow model. In all cases, we train for 20K steps using the Adam optimiser with $10^{-3}$ learning rate, batch size 32, and 512 MC samples. Figure~\ref{fig:markovshapes-bpd} plots the performance of Flow-SSN against the baseline SSN, showing how Flow-SSN outperforms all SSN variants. Notably, Flow-SSN converges faster \textit{and} achieves a better final result than even the full-rank SSN. In Figure~\ref{fig:markovshapes-samples}, we show how samples from Flow-SSN faithfully represent each of the underlying shapes (i.e. \includegraphics[width=0.03\columnwidth]{images/square.png}, \includegraphics[width=0.03\columnwidth]{images/plus.png}, and \includegraphics[width=0.03\columnwidth]{images/dot.png}). In contrast, the rank 2 SSN introduces sampling errors, hallucinating nonexistent shapes due to poorly modelling the true covariance structure.

\begin{table*}[!t]
    \caption{\textbf{Quantitative results on LIDC-IDRI}. Flow-SSN achieves SOTA performance with fewer params. ($\Delta$/$\infty$ denote discrete/continuous time variants). Results with ($\dagger$) are from~\citet{zhang2022probabilistic}. $D^2_{\text{GED}}(M)$ denotes $16{<}M{\leq}100$ MC samples used. More baselines in App. E.}
    \centering
    \begin{tabular}{lcccccr}
    \toprule
     & \multicolumn{6}{c}{\textbf{LIDC-IDRI} (128${\times}$128)} \\[2pt]
    \textsc{Method} & $D^2_{\text{GED}} (16) \downarrow$ & $D^2_{\text{GED}}(M) \downarrow$ & Diversity $\uparrow$ & 
    Dice $\uparrow$ & HM-IoU $\uparrow$ & \#Param
    \\
    \midrule
    UNet~\cite{ronneberger2015u}$^{\dagger}$ & - & 0.676{\scriptsize $\pm$.000}
    & - & 
    0.519{\scriptsize $\pm$.004} 
    & 0.463{\scriptsize $\pm$.000} & 9M \\
    ProbUNet~\cite{kohl2018probabilistic} & 0.287{\scriptsize $\pm$N/A} & 0.252{\scriptsize $\pm$.004} & 0.469{\scriptsize $\pm$.003} &  
    0.390{\scriptsize $\pm$.004} 
    & 0.500{\scriptsize $\pm$.030} & 18M \\    cFlow~\cite{selvan2020uncertainty}$^{\dagger}$ & - & 0.225{\scriptsize $\pm$.002} & 0.523{\scriptsize $\pm$.010} & 0.449{\scriptsize $\pm$.000} &  0.584{\scriptsize$\pm$.000} & N/A \\
    PHiSeg~\cite{baumgartner2019phiseg} & - & 0.225{\scriptsize $\pm$.004} & 0.496{\scriptsize $\pm$.003} & 0.486{\scriptsize $\pm$.010} & 0.595{\scriptsize$\pm$.00$^{\dagger}$} & 63M \\
    SSN~\cite{monteiro2020stochastic} & - & 0.225{\scriptsize $\pm$.002} & 0.609{\scriptsize $\pm$.002} & 0.436{\scriptsize $\pm$.004} & 0.555{\scriptsize$\pm$.01$^{\dagger}$} & 41M \\
    P$^2$SAM~\cite{huang2024p2sam}& 0.218{\scriptsize $\pm$N/A} & 0.216{\scriptsize $\pm$N/A} & - & - & 0.679{\scriptsize $\pm$N/A} & N/A \\
    SSN++ (ours) & 0.241{\scriptsize $\pm$.001} & 0.212{\scriptsize $\pm$.001} & 0.575{\scriptsize $\pm$.005} & 0.471{\scriptsize $\pm$.003} & - & 20M \\
    JProbUNet~\cite{zhang2022probabilistic} & - & 0.206{\scriptsize $\pm$.000} & 0.475{\scriptsize $\pm$.010} & 0.511{\scriptsize $\pm$.010} & 0.647{\scriptsize $\pm$.010} & N/A \\
    MoSE~\cite{gao2023modeling}& 0.218{\scriptsize $\pm$.003} & 0.189{\scriptsize $\pm$.002} & - & - & 0.624{\scriptsize $\pm$.004} & 42M \\
    \midrule
    Flow-SSN$_{\Delta}$ \hspace{1.25pt}: {\footnotesize $\{\text{IAF, 1-step}\}$} & 0.240{\scriptsize $\pm$.002} & 0.212{\scriptsize $\pm$.001} & 0.598{\scriptsize $\pm$.000} & 0.468{\scriptsize $\pm$.002} & \textbf{0.879}{\scriptsize $\pm$.000} & {14M}
    \\
    Flow-SSN$_{\infty}$ : {\footnotesize $\{\text{Dopri5}, \text{tol}{=}\text{1e{-}6}\}$} & 0.209{\scriptsize $\pm$.002} & 0.182{\scriptsize $\pm$.001} & 0.521{\scriptsize $\pm$.005} & 0.610{\scriptsize $\pm$.003} & 0.872{\scriptsize $\pm$.000} & {14M} 
    \\
    Flow-SSN$_{\infty}$ : {\footnotesize $\{\text{Euler}, T{=}50\}$} & 0.210{\scriptsize $\pm$.002} & 0.182{\scriptsize $\pm$.000} & 0.518{\scriptsize $\pm$.006} & 0.611{\scriptsize $\pm$.003} & 0.873{\scriptsize $\pm$.000} & {14M} 
    \\
    Flow-SSN$_{\infty}$ : {\footnotesize $\{\text{Euler}, T{=}250\}$} & \textbf{0.207}{\scriptsize $\pm$.000} & \textbf{0.181}{\scriptsize $\pm$.000} & 0.520{\scriptsize $\pm$.006} & \textbf{0.611}{\scriptsize $\pm$.007} & 0.873{\scriptsize $\pm$.001} & {14M} 
    \\
    \bottomrule
    \end{tabular}
    \label{tab:lidc_table}
\end{table*}
\paragraph{Rank Underspecification Ratio.} The underspecification ratio used for the SSN models on \textit{MarkovShapes} (most extreme is $r{=}2 : r_{\text{true}}{=}12$) is mild compared to more complex high-dimensional real-world data, where $r \ll r_{\text{true}}$. Thus, the low-rank assumption of SSNs, typically $r{=}10$, is likely even more punishing in real-world settings. As shown in the next experiment, simply increasing the assumed rank for real-world data collapses the SSN to a deterministic model.
\subsection{Lung Nodule Segmentation}
\label{subsec: LIDC}
The LIDC-IDRI dataset~\cite{armato2011lung} is a standard benchmark for evaluating stochastic segmentation models, with multiple ground truth label maps per image reflecting the inherent variability among medical experts. LIDC contains 1018 lung CT scans, with lung nodule annotation masks provided by 4 radiologists (from a pool of 12). We use the preprocessing from~\citet{kohl2018probabilistic,baumgartner2019phiseg}, where  $128\times128$ patches are extracted such that each patch is centred on a nodule. The process yields 15,096 slices with 4 segmentations each, with a $60/20/20$ train/valid/test split. As in previous work, we measure performance using Generalised Energy Distance (GED), Dice, and Hungarian-Matched Intersection over Union (HM-IoU)~\cite{kohl2019hierarchical,rahman2023ambiguous,zbinden2023stochastic}. Our architecture is a streamlined version of the~\citet{dhariwal2021diffusion}'s UNet to parameterise both the prior and the continuous-time flow. We use a \textit{single} autoregressive Transformer layer to parameterise our discrete-time Flow-SSN; the IAF variant trained using Eq.~\eqref{eq:discrete_xlike}. For more details/results, see Appendix D/E.

\begin{figure}[t]
    \centering
    \includegraphics[clip, trim={15 0 15 0}, width=\columnwidth]{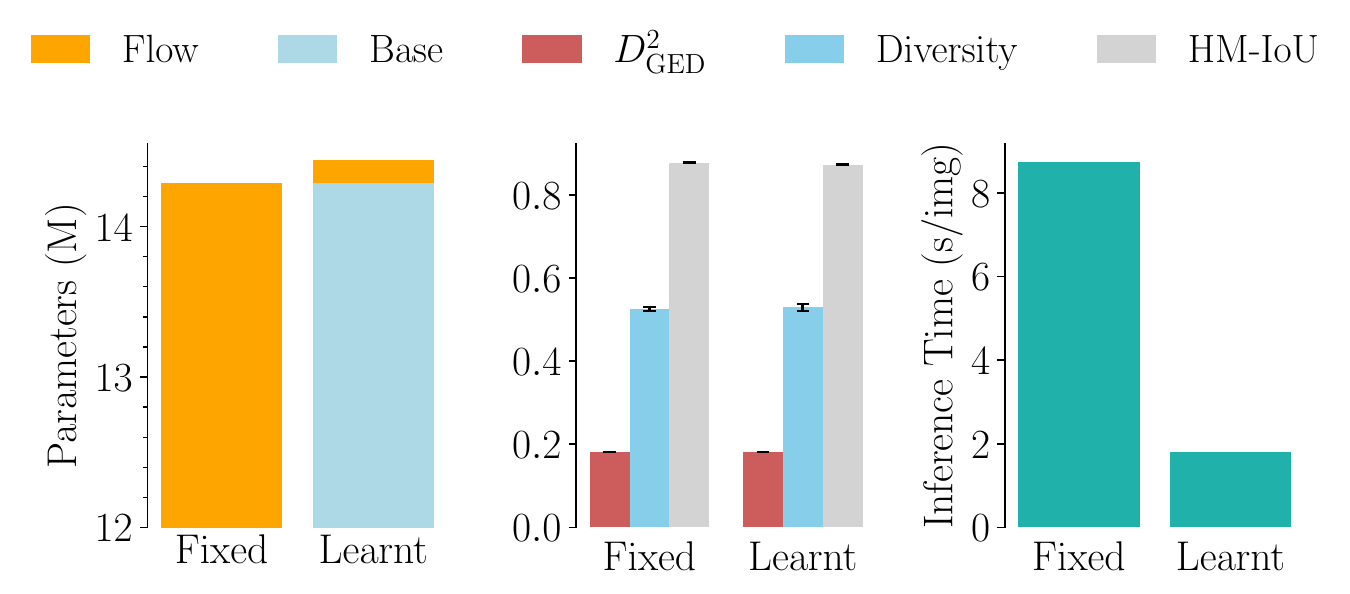}
    \caption{\textbf{Comparing learnt vs fixed priors in Flow-SSNs}. We observe a $5\times$ reduction in inference time without sacrificing performance. The flow network used in the learnt prior setup is around $100\times$ smaller than the fixed prior baseline, making ODE solving much cheaper. Note that $\mathrm{s/img}$ here includes inferring 100 MC samples per image, effectively performing 100 forward passes.}
    \label{fig:flowssn_eff}
\end{figure}

\begin{figure}[t]
    \centering
    \includegraphics[clip, trim={0 0 0 0}, width=\columnwidth]{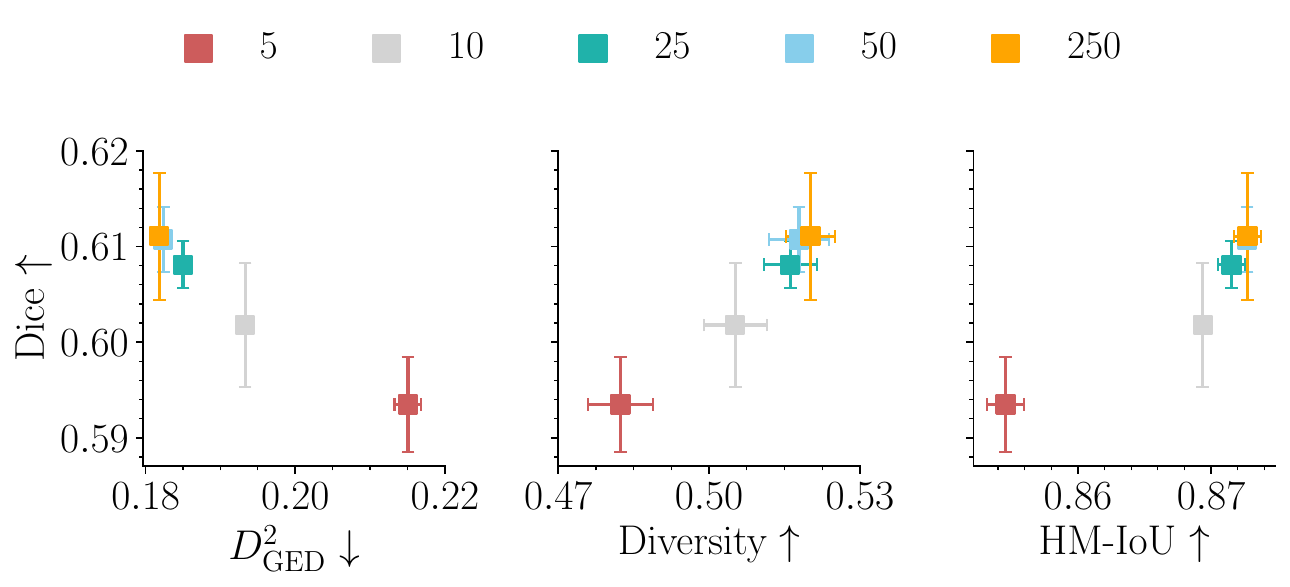}
    \caption{\textbf{Ablation study of the number of ODE solving steps}. Results (LIDC-IDRI) obtained using the Euler method. Flow-SSNs can perform competitively with as few as $T{=}10$ ODE solver steps.}
    \label{fig:flowssn_T_ablation}
\end{figure}

Table~\ref{tab:lidc_table} reports our results. Our baseline SSN (SSN++) outperforms the original~\cite{monteiro2020stochastic}, and the Flow-SSN achieves state-of-the-art performance in all metrics using fewer parameters\footnote{We note that diversity is only relevant contextually, as high diversity can be trivially achieved with random noise as a model.}. Fig.~\ref{fig:flowssn_eff} shows ablation results for learning the base distribution (i.e. prior) vs holding it fixed (as typically done). We observe a \textit{significant} reduction in inference time without sacrificing performance, thanks to using a small network for the flow and allocating most of the model capacity to the prior (150K vs 14M params). Flow-SSN is ${\simeq}10\times$ more efficient than CCDM~\cite{zbinden2023stochastic} (cf. App. E), and can perform competitively with just 10 ODE solver steps (c.f. Fig.~\ref{fig:flowssn_T_ablation}).
\begin{figure*}[!t]
    \centering
    \includegraphics[clip, trim={0 0 0 0}, width=.99\textwidth]{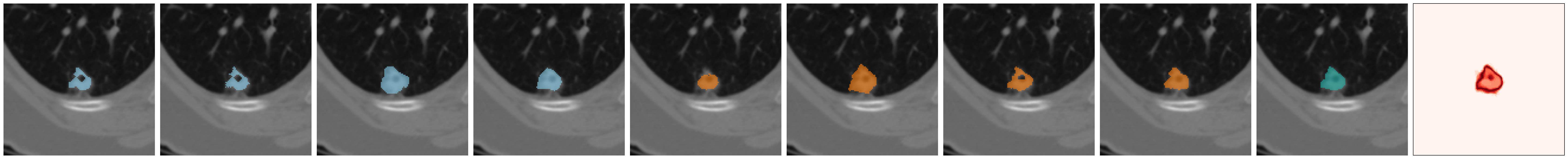}
    \\[-1.5pt]
    \includegraphics[clip, trim={0 0 0 0}, width=.99\textwidth]{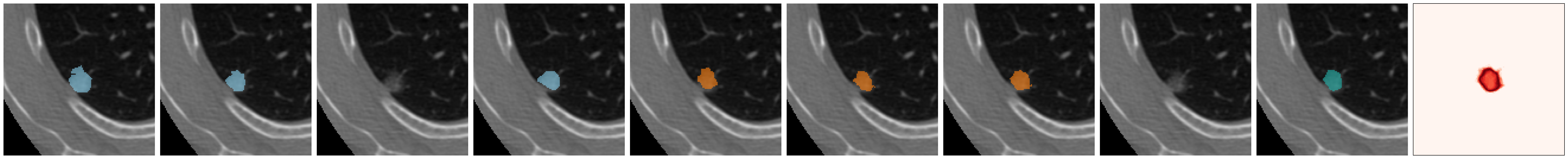}
    \\[-1.5pt]
    \includegraphics[clip, trim={0 0 0 0}, width=.99\textwidth]{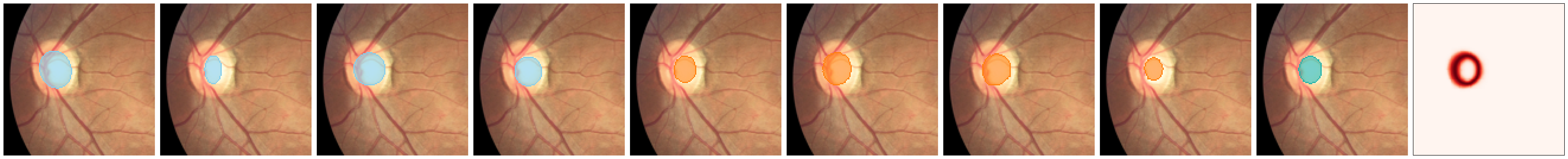}
    \\[-1.5pt]
    \includegraphics[clip, trim={0 0 0 0}, width=.99\textwidth]{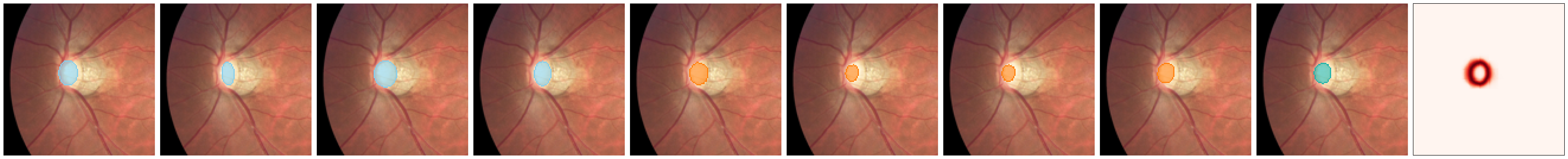}
    \caption{\textbf{Qualitative results on LIDC-IDRI (\textit{Rows 1, 2}) and REFUGE-MultiRater (\textit{Rows 3, 4}) using Flow-SSN}. (\textit{Cols. 1-4}) Multiple ground-truth expert segmentations; (\textit{Cols. 5-8}) Non-cherry-picked model samples; (\textit{Cols. 9, 10}) Mean prediction and pixel uncertainty map.}
    \label{fig:lidc_samples}
\end{figure*}
\begin{table}[!t]
    \small
    \caption{\textbf{Stochastic segmentation on REFUGE-MultiRater}. We set \textit{new} benchmarks for $D^2_{\text{GED}}$, Diversity and HM-IoU. We use $\Delta$ and $\infty$ to denote discrete and continuous-time Flow-SSN variants.}
    \centering
    \begin{tabular}{lcccc}
    \toprule
      & & \multicolumn{3}{c}{\textbf{REFUGE-MultiRater} (256${\times}$256)} \\[3pt]
    \textsc{Method} & $M$ & $D^2_{\text{GED}} \downarrow$ & Diversity $\uparrow$  & HM-IoU $\uparrow$
    \\
    \midrule
    \multirow{3}{*}{Flow-SSN$_{\Delta}$} & 16 & 0.116{\scriptsize $\pm$.007} & 0.441{\scriptsize $\pm$.012} & 0.766{\scriptsize $\pm$.006}
    \\
    & 100 & 0.089{\scriptsize $\pm$.004} & 0.462{\scriptsize $\pm$.013} & 0.851{\scriptsize $\pm$.001}
    \\
    & 512 & \textbf{0.081}{\scriptsize $\pm$N/A} & 0.447{\scriptsize $\pm$N/A} & \textbf{0.881}{\scriptsize $\pm$N/A}
    \\
    \midrule
    \multirow{2}{*}{Flow-SSN$_{\infty}$} & 16 & 0.112{\scriptsize $\pm$.003} & 0.424{\scriptsize $\pm$.004} & 0.751{\scriptsize $\pm$.003}
    \\
    & 100 & \textbf{0.089}{\scriptsize $\pm$.001} & 0.453{\scriptsize $\pm$.020} & \textbf{0.832}{\scriptsize $\pm$.004}
    \\
    \bottomrule
    \end{tabular}
    \label{tab:ref_table}
\end{table}
\begin{figure}[!t]
    \centering
    \includegraphics[clip, trim={0 0 0 0}, width=\columnwidth]{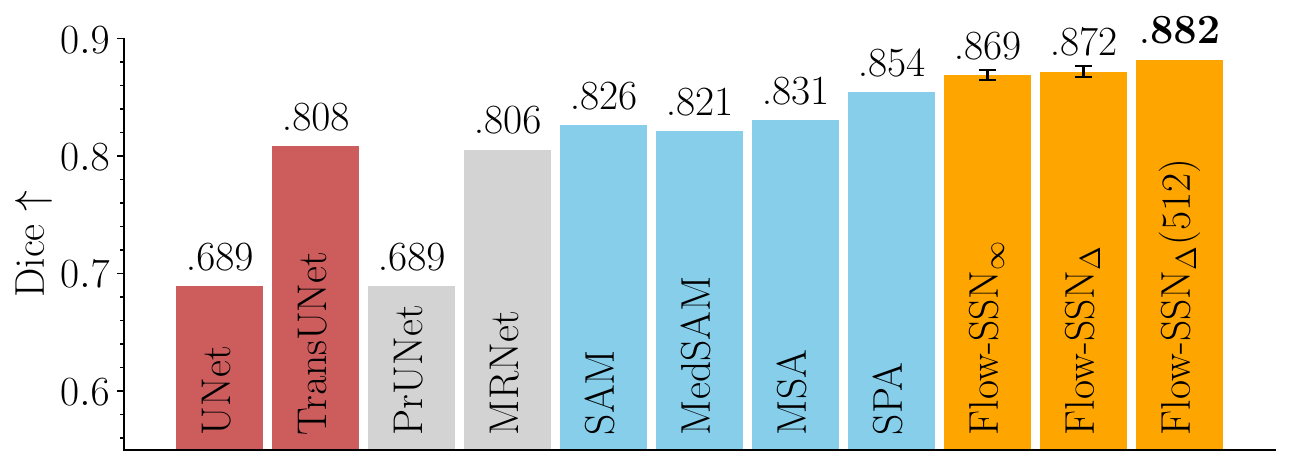}
    \caption{\textbf{Optical cup segmentation on REFUGE-MultiRater}. Baselines are from~\citet{zhu2024spa}. We observe impressive Dice performance using modestly-sized Flow-SSNs with only 15M params.}
    \label{fig:flowssn_refuge_results}
\end{figure}
\subsection{Optical Cup Segmentation}
\label{subsec: refuge}
REFUGE2~\cite{fang2022refuge2} is a publicly available benchmark dataset for optic disk and optic cup segmentation. Each fundus image in the dataset has multiple associated ground-truth label maps. Specifically, it includes annotations from 7 different independent ophthalmologists, each with an average of 8 years of experience. It contains a total of 1200 images, 400 for training, validation, and testing, respectively. Segmentation of the optic cup is inherently more variable than segmentation of the optic disk~\cite{fang2022refuge2}, therefore, we only consider the former in this work. As before, we measure performance using GED, Dice, and HM-IoU. For training, we resize the images to 256$\times$256 to match the output size used by~\citet{zhu2024spa} and enable fair comparisons with their reported Dice baselines. Namely, TransUNet~\cite{chen2021transunet}, MRNet~\cite{chen2021transunet}, SAM~\cite{kirillov2023segment}, MedSAM~\cite{ma2024segment} and MSA \cite{zhang2024segment}. For training details, please refer to Appendix D,
and for extra ablation results see Appendix E.
As shown in Fig.~\ref{fig:flowssn_refuge_results}, Flow-SSNs outperform all baselines by a considerable margin using only 15M parameters. Table~\ref{tab:ref_table} provides \textit{new} stochastic segmentation benchmark results, featuring both discrete and continuous-time Flow-SSN variants.

Surprisingly, we find that a discrete-time autoregressive Flow-SSN (a single-step IAF trained as per Eq.~\eqref{eq:discrete_xlike}) outperforms the continuous-time one on this dataset. We expect continuous-time Flow-SSNs to outperform shallow discrete-time ones in general, but this result encourages further exploration into this model class, as the discrete-time Flow-SSN is ${\approx}200\times$ faster to sample from than the continuous-time one. For this reason, we were able to push the number of MC samples used for evaluation up to $M{=}512$ and observed further significant gains in performance on all metrics.
\section{Conclusion}
\label{sec: conclusion}
We introduce the Flow Stochastic Segmentation Network (Flow-SSN), a generative segmentation model featuring both discrete-time autoregressive and modern continuous-time flow variants. By overcoming the constraints of low-rank parameterisations in standard SSNs, Flow-SSNs enable the estimation of arbitrarily high-rank pixel-wise covariances without requiring prior assumptions about rank or explicit storage of distributional parameters. Moreover, Flow-SSNs significantly improve sampling efficiency compared to standard diffusion-based segmentation models, thanks to a key architectural design that includes learning the base distribution (i.e. the prior) of the flow, which is typically fixed. Experimental results on standard real-world benchmarks demonstrate the efficacy of Flow-SSNs, achieving state-of-the-art performance, and highlighting their potential to advance stochastic segmentation in safety-critical applications with inherent ambiguities, such as medical imaging.

\newpage
\section*{Acknowledgments}
F.R. and B.G. acknowledge the support of the UKRI AI programme, and the EPSRC, for CHAI-EPSRC Causality in Healthcare AI Hub (grant no. EP/Y028856/1). O.T. and R.M. are funded by the European Union’s Horizon Europe research and innovation programme under grant agreement 101080302. C.J. is
supported by Microsoft Research, EPSRC, and The Alan
Turing Institute through a Microsoft PhD scholarship and
a Turing PhD enrichment award. A.K. is supported by UKRI (grant no. EP/S023356/1), as
part of the UKRI Centre for Doctoral Training in Safe \&
Trusted AI, and acknowledges support from the EPSRC Doctoral Prize. B.G. received
support from the Royal Academy of Engineering as part of his Kheiron/RAEng Research Chair.
{
    \small
    \bibliographystyle{ieeenat_fullname}
    \bibliography{bibfile.bib}
}
\newpage
\makeatletter
\def\iccvsect#1{\iccvsection{#1}}%
\def\iccvsubsect#1{\iccvsubsection{#1}}%
\def\iccvsubsubsect#1{\iccvsubsubsection{#1}}%
\makeatother
\appendix
\onecolumn
\addcontentsline{toc}{section}{Appendices}
\part{Appendices}
{
    \changelinkcolor{black}{}
    \parttoc
}
\section{Theoretical Rank Analysis: Proofs}
\label{app: proofs}
\begin{lemma}[Rank Increase] Let the logits $\boldsymbol{\eta} \in \mathbb{R}^{kd}$ be low-rank Gaussian distributed $\boldsymbol{\eta} \mid \mathbf{x} \sim \mathcal{N}(\boldsymbol{\mu}(\mathbf{x}), \boldsymbol{\Sigma}(\mathbf{x})) = p_{H|X}$, where the covariance matrix $\boldsymbol{\Sigma}(\mathbf{x}) \in \mathbb{R}^{kd \times kd}$ has rank $\mathrm{rank}(\boldsymbol{\Sigma}(\mathbf{x})) = r$. If we define $\mathbf{y} = g(\boldsymbol{\eta})$, where $g = \mathrm{softmax}_k$, the rank of the covariance matrix $\mathrm{Cov}(\mathbf{y})$ of the pushforward distribution $g_\# p_{H|X}$ increases under the condition:
\begin{align}
    \mathrm{rank}(
    \mathrm{Cov}(\mathbf{y})) 
    > r \iff r < d(k-1).
\end{align}
\begin{proof}
    First recall that we use $\mathrm{softmax}_k(\cdot)$ to denote that the softmax function is applied row-wise across the $k$ dimension (i.e. over the number of categories $k$), after reshaping the input logits $\boldsymbol{\eta} \in \mathbb{R}^{kd}$ to $\boldsymbol{\eta} \in \mathbb{R}^{k\times d}$. More formally:
    \begin{align}
        &&\mathbf{y}_{:, j} = \mathrm{softmax}(\boldsymbol{\eta}_{:,j}), && \mathrm{where} && y_{ij} = \frac{e^{\eta_{ij}}}{\sum_{l=1}^k e^{\eta_{lj}}}, && \mathrm{for} \quad i=1,\dots,k, \ j=1,\dots,d. &&
    \end{align}
    
    Now let $\mathbf{f} : \mathbb{R}^{kd} \to (0,1)^{kd}$ be the composition of $\mathrm{softmax}_k(\boldsymbol{\eta} )$ then flattening $\mathbf{y} \in (0,1)^{k\times d}$ back to $\mathbf{y} \in (0, 1)^{kd}$. The resulting Jacobian $\mathbf{J}_\mathbf{f} \in \mathbb{R}^{kd \times kd}$ of $\mathbf{f}$ is a sparse matrix given by:
    \begin{align}
    &&\mathbf{J}_{\mathbf{f}} = 
    \begin{bmatrix}
        \frac{\partial y_1}{\partial \eta_1} & \cdots & \frac{\partial y_1}{\partial \eta_{kd}}\\[5pt]
        \vdots                             & \ddots & \vdots\\[5pt]
        \frac{\partial y_{kd}}{\partial \eta_1} & \cdots & \frac{\partial y_{kd}}{\partial \eta_{kd}}
    \end{bmatrix},
    &&
    \mathrm{where}
    &&
    \frac{\partial y_i }{\partial \eta_j} \neq 0 \quad \mathrm{if} \quad j \equiv i \pmod k.
    && 
    \end{align}
    In words, $\mathbf{J}_{\mathbf{f}}$ has a total of $dk^2$ non-zero entries, including the diagonal and a checkerboard-like pattern with column-wise intervals of stride $k$, caused by the flattening operation.
    
    Intuitively, $\mathbf{f}$ non-linearly couples each element in $\boldsymbol{\eta}$ with $k-1$ number of other elements in $\boldsymbol{\eta}$ to produce $\mathbf{y}$, i.e. each $\boldsymbol{\eta}_{:,j} \in \mathbb{R}^{k \times 1}$ is independently mapped to the $(k-1)$-dimensional simplex $\Delta^{k-1}$ via an element-wise softmax, for $j=1,2,\dots,d$. As a result, we can conclude that $\mathrm{rank}(\mathbf{J}_{\mathbf{f}}) = d(k-1)$.
    
    Considering a Taylor expansion of $\mathrm{Cov}(\mathbf{y})$, we know that the first-order term only contributes linearly to the result:
    \begin{align}
        &&\mathrm{Cov}(\mathbf{y}) = \mathbf{J}_{\mathbf{f}}\boldsymbol{\Sigma}(\mathbf{x}) \mathbf{J}_{\mathbf{f}}^\top + \cdots, && \mathrm{and} && \mathrm{rank}(\mathbf{J}_{\mathbf{f}}\boldsymbol{\Sigma}(\mathbf{x}) \mathbf{J}_{\mathbf{f}}^\top) = \mathrm{min}(r, d(k-1)) = r.&&
    \end{align}
    Since $\mathbf{f}$ is non-linear, the higher-order terms in the Taylor expansion (and their rank) must be non-zero. Thus, by the rank subadditivity property: $\mathrm{rank}(A + B) \leq \mathrm{rank}(A) + \mathrm{rank}(B)$, we conclude that $\mathrm{rank}(\mathrm{Cov}(\mathbf{y})) > r$ iff $r < d(k-1)$.
\end{proof}
\end{lemma}

\begin{theorem}[Sublinear Growth of the Effective Rank]
\label{prop: sublin}
Given a low-rank Gaussian covariance matrix $\boldsymbol{\Sigma}(\mathbf{x}) \in \mathbb{R}^{kd \times kd}$ with initial rank $\mathrm{rank}(\boldsymbol{\Sigma}(\mathbf{x})) = r < d(k-1)$. The increase in the effective rank $\mathrm{erank}(\mathrm{Cov}(\mathbf{y}))$, in the sense of Lemma~\ref{prop: rank_increase}, grows sublinearly with respect to the initial rank $r$.

\begin{proof} Recall that the rank of $\boldsymbol{\Sigma}(\mathbf{x}) \in \mathbb{R}^{kd \times kd}$ is by definition given by:
\begin{align}
    &&\mathrm{rank}(\boldsymbol{\Sigma}(\mathbf{x})) = \sum_{i=1}^{kd} \mathbb{I}(\sigma_i > 0), && \mathrm{where} && \sigma_1 \geq \sigma_2 \geq, \dots, \sigma_{kd} \geq 0 &&
\end{align}
are the singular values of $\boldsymbol{\Sigma}(\mathbf{x})$, and $\mathbb{I}(\cdot)$ is the indicator function. Since $\mathrm{rank}(\boldsymbol{\Sigma}(\mathbf{x})) = r$ we have $\sigma_1 \geq, \dots, \geq \sigma_{r} > 0$.

Given that the non-linear pushforward operation defined in Lemma~\ref{prop: rank_increase} is rank-increasing (as long as $r < d(k-1)$), we can assert that the final rank, $\mathrm{rank}(\mathrm{Cov}(\mathbf{y})) = r_1$, grows \textit{at least linearly} w.r.t. the initial rank $\mathrm{rank}(\boldsymbol{\Sigma}(\mathbf{x})) = r$. This is intuitive as each non-zero singular value of $\boldsymbol{\Sigma}(\mathbf{x})$ contributes at least one additional rank increment to $\mathrm{rank}(\mathrm{Cov}(\mathbf{y}))$.

This holds for the \textit{effective rank} $\mathrm{erank}(\mathrm{Cov}(\mathbf{y}))$ if and only if the distribution of singular values of $\mathrm{Cov}(\mathbf{y})$, $p$, is uniform:
\begin{align}
    \label{eq: eff_rank_grow}
    &&\mathrm{erank}(\mathrm{Cov}(\mathbf{y})) = \mathrm{rank}(\mathrm{Cov}(\mathbf{y})) \iff p_i = \frac{1}{r_1}, \quad \mathrm{for} \quad i=1,2,\dots,r_1, &&
\end{align}
which is straightforward to show by substituting $p$ into the definition of effective rank (c.f. Definition~\ref{def: effective_rank}):
\begin{align}   
    H(p) = -\sum_{i=1}^{r_1}\frac{1}{r_1} \log \left(\frac{1}{r_1}\right) = \log r_1 \implies \mathrm{erank} 
    (\mathrm{Cov}(\mathbf{y})) = e^{H(p)} = r_1 = \mathrm{rank}(\mathrm{Cov}(\mathbf{y})).
\end{align}

Alternatively, when the singular value distribution $p$ is non-uniform, e.g. skewed, we have that:
\begin{align}
    \sigma_1 \gg \sigma_2 \geq,\dots, \geq \sigma_{r_1} > 0 \implies H(p) < \log r_1 
    \implies \mathrm{erank}(\mathrm{Cov}(\mathbf{y}))  < \mathrm{rank}(\mathrm{Cov}(\mathbf{y}))
    ,
\end{align}
since the uniform is the maximum entropy distribution and $H(p)$ is bounded above by $\log r_1$. For a random matrix with continuous entries and $kd>1$, the probability that all its singular values are equal is zero; so the singular value distribution $p$ is almost surely non-uniform. Thus, since linear growth holds iff $p$ is uniformly distributed (c.f. Eq.~\eqref{eq: eff_rank_grow}), the final effective rank $\mathrm{erank}(\mathrm{Cov}(\mathbf{y})) = r_1$ must grow \textit{sublinearly} w.r.t. the initial rank $\mathrm{rank}(\boldsymbol{\Sigma}(\mathbf{x})) = r$, concluding the proof.
\end{proof}
\begin{remark}
    One way to intuit this result is by recalling that entropy grows logarithmically with the number of dimensions in uniform distributions. For non-uniform distributions, the growth of entropy can be slower, depending on how the additional event (represented by, say, $\sigma_{r+1} > 0$ in our case) relates to the existing probabilities $p_1, p_2,\dots, p_{r}$.
\end{remark}
\end{theorem}

\section{Flow Stochastic Segmentation Networks: Proofs}
\label{app: discrete flow}
We begin by reviewing autoregressive image models, then proceed to theoretical results and design options for Flow-SSNs. 
\subsection{Autoregressive Image Models}
Autoregressive models factorise joint probability distributions into a product of conditional distributions using the chain rule of probability. 
For example, by representing an image $\mathbf{y}$ as a sequence of pixels $y_1,y_2,\dots,y_T$, we can estimate the joint pixel distribution by maximum likelihood estimation on the parameters of a model:
\begin{align}
    && \hat{\boldsymbol{\theta}}_{\text{MLE}} = \argmax_{\boldsymbol{\theta}} \sum_{i=1}^N \log p(\mathbf{y}_i;\boldsymbol{\theta}),
    &&
    p(\mathbf{y}_i;\boldsymbol{\theta}) = \prod_{t=1}^T p(y_{i,t} \mid y_{i,1},\dots,y_{i,t-1};\boldsymbol{\theta}_t).&&
\end{align}
Autoregressive image models~\citep{van2016conditional,van2016pixel, salimans2017pixelcnn} can estimate arbitrary joint pixel distributions but are computationally costly to sample from as each pixel must be generated sequentially. Discrete-time autoregressive Flow-SSNs avoid sequential sampling.
\begin{proposition}[Full Covariance Flow Transformation]
Let $\mathbf{u} = (u_1, u_2, \dots, u_d)^{\top}$ be a Gaussian random vector with diagonal covariance $\mathcal{N}(\boldsymbol{\mu}, \mathrm{diag}(\boldsymbol{\sigma}^2))$. A linear autoregressive model is sufficient to transform $\mathbf{u}$ into a new variable $\boldsymbol{\eta} \in \mathbb{R}^d$ with full covariance $\boldsymbol{\Sigma} \in \mathbb{R}^{d \times d}$.
\begin{proof}
    First recall that autoregressive models can represent any joint distribution as a product of conditionals:
        $p(\boldsymbol{\eta}; \theta) = \prod_{i=1}^d p(\eta_i \mid \eta_{1}, \dots ,\eta_{i-1}; \theta).$
    Define the autoregressive transformation of $\mathbf{u}$ into $\boldsymbol{\eta}$ as follows:
   \begin{align}
       &&\eta_i = \mu_i(\boldsymbol{\eta}_{1:i-1}) + \sigma_i(\boldsymbol{\eta}_{1:i-1})u_i, && \mathrm{for} \quad i=1,2,\dots,d, &&
   \end{align}
    where $\mu_i(\cdot)$ and $\sigma_i(\cdot)$ are functions of preceding elements $\boldsymbol{\eta}_{1:i-1}$, thereby inducing a lower triangular dependency structure in $\boldsymbol{\eta}$. Thus, there exists a lower triangular matrix $\mathbf{L} \in \mathbb{R}^{d\times d}$ such that
   $\mathbf{L}\mathbf{L}^{\top} = \boldsymbol{\Sigma}$ (via Cholesky decomposition), and the autoregressive transformation can be equivalently expressed as a set of linear equations of the form:
   \begin{align}
        &&\eta_i = \sum_{j=1}^{i-1} L_{ij}u_j + L_{ii}u_i, && \mathrm{for} \quad i=1,2, \dots, d, &&
   \end{align}
   where $L_{ij}$ determine the linear dependencies between elements, thereby inducing a full covariance structure for $\boldsymbol{\eta}$.
\end{proof}
\begin{remark}
    \citet{kingma2016improved} undoubtedly recognised this fact, as it was informally mentioned in their exposition. We provide a simple proof here to make the argument in favour of our proposed approach more rigorous.
\end{remark}
\end{proposition}

\subsection{Choosing a Flow \& Optimization Objective}
\label{app: choosing a flow}
The following provides supplementary derivations of the different optimization objectives for training discrete-time Flow-SSNs outlined in the main paper. Recall the discrete-time Flow-SSN is defined as follows:
\begin{align}
    p(\mathbf{y} \mid \mathbf{x}) &= \int p(\mathbf{y} \mid \boldsymbol{\eta}) p(\boldsymbol{\eta} \mid \mathbf{x};\lambda,\theta) \ \mathrm{d}\boldsymbol{\eta} \label{app:like}
    \\[3pt]
    \qquad \mathrm{where} \quad p(\boldsymbol{\eta} \mid \mathbf{x};\lambda, \theta) = p_{U|X}(\mathbf{u} \mid \mathbf{x}; \lambda) &\left|\det \mathbf{J}_{\phi}(\mathbf{u})\right|^{-1},
    \quad \mathrm{and} \quad p(\mathbf{y} \mid \boldsymbol{\eta}) = \mathrm{Cat}(\mathbf{y};\mathrm{softmax}_k(\boldsymbol{\eta})). \qquad
\end{align}
We then have to choose a flow to model $p(\boldsymbol{\eta} \mid \mathbf{x};\lambda, \theta)$, and for this we revisit affine autoregressive flows, specifically IAFs~\cite{kingma2016improved} and MAFs~\cite{papamakarios2017masked} as they are both simple and flexible enough for our needs. They remain relatively underexplored to date, and in this work, we combine them with modern autoregressive Transformers to build Flow-SSNs. With that in mind, in the following subsections, we detail the various design options available for Flow-SSNs.

\subsection{Expected Categorical Likelihood: Monte Carlo Estimator} 
The simplest approach is to use an IAF $p^{\text{IAF}}(\boldsymbol{\eta} \mid \mathbf{x}; \lambda, \theta)$, which is fast to sample form, and use a simple Monte Carlo estimator of the likelihood in Eq.~\eqref{app:like} analogous to a standard SSN:
\begin{align}
    p(\mathbf{y} \mid \mathbf{x}) &= \mathbb{E}_{\boldsymbol{\eta} \sim p^{\text{IAF}}(\boldsymbol{\eta} \mid \mathbf{x}; \lambda, \theta)} \left[ p(\mathbf{y} \mid \boldsymbol{\eta}) \right] 
    \\[2pt] & = \mathbb{E}_{\mathbf{u} \sim p_{U|X}(\mathbf{u} \mid \mathbf{x}; \lambda)} \left[ p(\mathbf{y} \mid \boldsymbol{\eta} = \phi(\mathbf{u};\theta)) \right]
    \\[2pt] & \approx \frac{1}{M}\sum_{i=1}^M p(\mathbf{y} \mid \phi(\mathbf{u}^{(i)};\theta)), \qquad \mathbf{u}^{(i)}|\mathbf{x} \sim p_{U|X},
\end{align}
where $(\mathbf{x}, \mathbf{y}) \sim p_{\text{data}}(\mathbf{x}, \mathbf{y})$ and the flow is parameterised by an autoregressive model with parameters $\theta$. Taking logs, we get:
\begin{align}
    \log p(\mathbf{y} \mid \mathbf{x}) &\approx \log \frac{1}{M}\sum_{i=1}^M p(\mathbf{y} \mid \phi(\mathbf{u}^{(i)};\theta))
    \\[2pt] & = \mathrm{LSE} \left(\log p(\mathbf{y} \mid \phi(\mathbf{u}^{(1)};\theta) ) + \cdots + \log p(\mathbf{y} \mid \phi(\mathbf{u}^{(M)};\theta))\right) - \log M,
\end{align}
where $\mathrm{LSE}(\cdot)$ is the log-sum-exp function, and the individual categorical likelihood terms are given by:
\begin{align}
    \log p(\mathbf{y} \mid \boldsymbol{\eta} = \phi(\mathbf{u};\theta)) = \log \mathrm{Cat}(\mathbf{y};\mathrm{softmax}_k(\boldsymbol{\eta})) = \sum_{i=1}^k \sum_{j=1}^d y_{i,j} \log \mathrm{softmax}(\boldsymbol{\eta}_{:,j})_i.
\end{align}
\begin{remark}
    This approach is the most similar to standard SSNs as it uses the same Monte Carlo setup to integrate out the logits and compute $p(\mathbf{y} \mid \mathbf{x})$. We simply replace the low-rank Gaussian parameterisation with an autoregressive flow that is still cheap to sample from: an IAF. The approach is attractive in that the majority of the model capacity is allocated to learning the base distribution $p_{U|X}$, which only requires a single forward pass to compute. Given the parameters of $p_{U|X}$, it is cheap to sample from it and compute the logits $\boldsymbol{\eta} = \phi(\mathbf{u};\theta)$, as the flow $\phi$ is lightweight, e.g. a single linear autoregressive transformation is sufficient (c.f. Proposition~\ref{prop: full_cov}). In practice, it can be beneficial to make $\phi$ more expressive.
\end{remark}
\subsection{Dual-Flow: Evidence Lower Bound}
An important fact about IAFs is that, although scoring observations is slow and unparallelizable, they can still score their \textit{own} samples efficiently since intermediate outputs can be cached when sampling $\boldsymbol{\eta} \sim p^{\text{IAF}}$, then reused for scoring the sample. This opens up various design options for Flow-SSNs, which we explain in greater detail next.

\subsubsection{Dual-Flow}
We introduce a \textit{dual-flow} setup comprised of an IAF $p^{\text{IAF}}(\boldsymbol{\eta} \mid \mathbf{x}; \lambda, \theta)$ and an MAF $p^{\text{MAF}}(\boldsymbol{\eta} \mid \mathbf{x}; \hat{\lambda}, \hat{\theta})$, both defined in logit space and trained concurrently, such that we maximize a lower bound on $\log p(\mathbf{y} \mid \mathbf{x})$:
\begin{align}
    \log p(\mathbf{y} \mid \mathbf{x}) &= \log \int p(\mathbf{y} \mid \boldsymbol{\eta}) p^{\text{MAF}}(\boldsymbol{\eta} \mid \mathbf{x};\hat{\lambda},\hat{\theta}) \ \mathrm{d}\boldsymbol{\eta}
    \\[2pt] &= \int p(\mathbf{y} \mid \boldsymbol{\eta})p^{\text{MAF}}(\boldsymbol{\eta} \mid \mathbf{x};\hat{\lambda},\hat{\theta})\frac{p^{\text{IAF}}(\boldsymbol{\eta} \mid \mathbf{x};\lambda,\theta)}{p^{\text{IAF}}(\boldsymbol{\eta} \mid \mathbf{x};\lambda,\theta)}\ \mathrm{d}\boldsymbol{\eta} 
    \\[2pt] & \geq
    \mathbb{E}_{\boldsymbol{\eta} \sim p^{\text{IAF}}(\boldsymbol{\eta} \mid \mathbf{x};\lambda,\theta)}\left[\log \frac{p(\mathbf{y} \mid \boldsymbol{\eta})p^{\text{MAF}}(\boldsymbol{\eta} \mid \mathbf{x};\hat{\lambda},\hat{\theta})}{p^{\text{IAF}}(\boldsymbol{\eta} \mid \mathbf{x};\lambda,\theta)}\right]
    \\[2pt] & =
    \mathbb{E}_{\boldsymbol{\eta} \sim p^{\text{IAF}}(\boldsymbol{\eta} \mid \mathbf{x};\lambda,\theta)}\left[\log p(\mathbf{y} \mid \boldsymbol{\eta})\right] - D_{\mathrm{KL}}\left(p^{\text{IAF}} \parallel p^{\text{MAF}}\right).
\end{align}
where $(\mathbf{x}, \mathbf{y}) \sim p_{\text{data}}(\mathbf{x}, \mathbf{y})$. There is no general closed-form solution to this KL term, so we approximate it using samples:
\begin{align}
    &&D_{\mathrm{KL}}\left(p^{\text{IAF}} \parallel p^{\text{MAF}}\right) \approx \frac{1}{M}\sum_{i=1}^M \log \frac{p^{\text{IAF}}(\boldsymbol{\eta}^{(i)} \mid \mathbf{x}; \lambda, \theta)}{p^{\text{MAF}}(\boldsymbol{\eta}^{(i)} \mid \mathbf{x}; \hat{\lambda}, \hat{\theta})}, && \boldsymbol{\eta}^{(i)}|\mathbf{x} \sim p^{\text{IAF}}, &&
\end{align}
which is cheap since sampling from $p^{\text{IAF}}$ is parallelizable and computing the base distribution $p_{U|X}$ only requires a single forward pass of $\mathbf{x}$. Furthermore, scoring $p^{\text{IAF}}$'s samples under $p^{\text{MAF}}$ is also cheap since scoring in MAFs is parallelizable.

In practice, we recommend using a different (unbiased) estimator which has lower variance, proposed by~\citet{schulman_kl_approximation}:
\begin{align}
    &&D_{\mathrm{KL}}\left(p^{\text{IAF}} \parallel p^{\text{MAF}}\right) \approx \frac{1}{M}\sum_{i=1}^M \mathrm{expm1}(r) - r, && r := \log\frac{p^{\text{IAF}}(\boldsymbol{\eta}^{(i)} \mid \mathbf{x}; \lambda, \theta)}{p^{\text{MAF}}(\boldsymbol{\eta}^{(i)} \mid \mathbf{x}; \hat{\lambda}, \hat{\theta})}, && \boldsymbol{\eta}^{(i)}|\mathbf{x} \sim p^{\text{IAF}}. &&
\end{align}
\begin{remark}
    This combination of IAFs and MAFs is reminiscent of \textit{probability density distillation}~\citep{oord2018parallel}, a student-teacher distillation technique used in audio synthesis models, wherein a pre-trained MAF is fixed as a teacher, and an IAF student learns to match it. In our setup, both flow models are trained concurrently to maximise a bespoke lower bound for stochastic segmentation tasks. Furthermore, it is possible for $p^{\text{IAF}}$ and $p^{\text{MAF}}$ to share the base distribution parameters $\lambda$.
\end{remark}

\subsubsection{Improper Uniform Prior} Alternatively to the above, we can replace $p^{\text{MAF}}$ with an improper uniform prior $p$ such that:
\begin{align}
    p(\boldsymbol{\eta}) = \mathrm{const}, \ \forall \boldsymbol{\eta} \implies D_{\mathrm{KL}}\left(p^{\mathrm{IAF}} \parallel p\right) &= 
    \mathbb{E}_{p^{\mathrm{IAF}}(\boldsymbol{\eta} \mid \mathbf{x};\lambda, \theta)}\left[ \log \frac{p^{\mathrm{IAF}}(\boldsymbol{\eta} \mid \mathbf{x};\lambda, \theta)}{\mathrm{const}}\right]
    \\ &= - H(p^{\mathrm{IAF}}) - \log \mathrm{const},
\end{align}
which, after dropping the constant term w.r.t the model parameters, yields the objective:
\begin{align}
    \log p(\mathbf{y} \mid \mathbf{x}) \geq
    \mathbb{E}_{\boldsymbol{\eta} \sim p^{\mathrm{IAF}}(\boldsymbol{\eta} \mid \mathbf{x};\lambda, \theta)}\left[\log p(\mathbf{y} \mid \boldsymbol{\eta})\right] + \beta H(p^{\mathrm{IAF}}),
\end{align}
with a weighting hyperparameter $\beta>0$. Maximising $H(p^{\mathrm{IAF}})$ prevents the model from collapsing to a deterministic one. The special case where $\beta$ is set to $0$ makes this objective equivalent to the expected categorical likelihood above.

\begin{proposition}[IAF Entropy Estimator] The entropy $H(p^{\mathrm{IAF}})$ can be efficiently estimated in parallel via Monte Carlo sampling $\mathbf{u}^{(i)}|\mathbf{x} \sim p_{U|X}$ using the following formula:
\begin{align}
     H(p^{\mathrm{IAF}}) \approx H(p_{U|X}) - \frac{1}{M} \sum_{i=1}^M \log \left|\det \mathbf{J}_{\phi^{-1}}
    (\phi(\mathbf{u}^{(i)} ; \theta))\right|, \qquad \mathbf{u}^{(i)}|\mathbf{x} \sim p_{U|X}.
\end{align}
\begin{proof} 
We simply start with the definition of differential entropy, then use a change-of-variables and simplify:
\begin{align}
    H(p^{\mathrm{IAF}}) &= -\mathbb{E}_{\boldsymbol{\eta} \sim p^{\text{IAF}}(\boldsymbol{\eta} \mid \mathbf{x};\lambda, \theta)}\left[ \log p^{\text{IAF}}(\boldsymbol{\eta} \mid \mathbf{x};\lambda, \theta) \right]
    \\[3pt] &= 
    -\int p^{\mathrm{IAF}}(\boldsymbol{\eta} \mid \mathbf{x};\lambda, \theta)\Big[ \log p_{U|X}(\phi^{-1}(\boldsymbol{\eta};\theta) \mid \mathbf{x};\lambda) + \log \left|\det \mathbf{J}_{\phi^{-1}}
    (\boldsymbol{\eta})\right| \Big] \ \mathrm{d} \mathbf{\boldsymbol{\eta}}
    \\[3pt] &= - \int p_{U|X}(\mathbf{u} \mid \mathbf{x};\lambda) \Big[\log p_{U|X}(\mathbf{u}\mid \mathbf{x};\lambda) + \log \left|\det \mathbf{J}_{\phi^{-1}}
    (\phi(\mathbf{u};\theta))\right| \Big] \ \mathrm{d} \mathbf{u}
    \\[3pt] & = H(p_{U|X}) - \mathbb{E}_{\mathbf{u} \sim p_{U|X}(\mathbf{u} \mid \mathbf{x};\lambda)} \left[ \log \left|\det \mathbf{J}_{\phi^{-1}}
    (\phi(\mathbf{u};\theta))\right| \right],
    \\[3pt] & \approx H(p_{U|X}) - \frac{1}{M} \sum_{i=1}^M \log \left|\det \mathbf{J}_{\phi^{-1}}
    (\phi(\mathbf{u}^{(i)};\theta))\right|, \qquad \mathbf{u}^{(i)}|\mathbf{x} \sim p_{U|X},
\end{align}
which is what we wanted to show.
\end{proof}
\begin{remark}
    The entropy $H(p_{U|X})$ is available in closed-form for typical base distributions (e.g. Gaussian).
    Computing $p_{U|X}(\mathbf{u} \mid \mathbf{x};\lambda)$ only requires a single forward pass, which is important as it comprises the majority of the model parameters. The flow component $\phi(\cdot)$ is lightweight, since a single (linear) layer is sufficient (c.f. Proposition~\ref{prop: full_cov}). Lastly, recall that autoregressive flows admit a lower triangular Jacobian by design, and as such, their log absolute determinant simplifies to a sum of the diagonal elements, which is easy to compute.
\end{remark}
\end{proposition}

\subsection{Logit Bijections}
For bijective logit mappings $\mathbf{y} = g(\boldsymbol{\eta})$ with tractable Jacobians determinants, the likelihood term $p(\mathbf{y} \mid \mathbf{x})$ in Flow-SSN models can be evaluated directly using the following change-of-variables formula:
\begin{align}
    p(\mathbf{y} \mid \mathbf{x}) = p_{U|X}((\phi \circ g)^{-1}(\mathbf{y}) \mid \mathbf{x}; \lambda) \nonumber
   \left|\det \mathbf{J}_{\phi^{-1}}(g^{-1}(\mathbf{y}))\right| \left|\det \mathbf{J}_{g^{-1}}(\mathbf{y})\right|.
\end{align}
However, the above does not hold when $g(\cdot)$ is the softmax function, as it is not bijective. TensorFlow Probability~\cite{abadi2016tensorflow} provides the somewhat underused function \texttt{tfp.bijectors.SoftmaxCentered}, which is an alternative bijective softmax transformation that we can use here, albeit at the cost of having to dequantise $\mathbf{y}$. We must also train with $k+1$ classes, where a newly introduced dummy class acts as a pivot and facilitates the bijective property of the function. Relatedly,~\citet{de2023high,monteiro2023measuring} have also recently used this transformation to train discrete causal mechanisms.
\section{Evaluation Metrics}
Throughout this work, we use the following evaluation metrics standard for stochastic segmentation tasks.
\paragraph{Dice Similarity Coefficient.}
For two label maps $\mathbf{y}, \hat{\mathbf{y}}$ of dimension $h \times w$, the Dice Similarity Coefficient (DSC) is defined as follows:
\begin{align}
      \text{DSC}(\mathbf{y},\hat{\mathbf{y}}) = \frac{2|\mathbf{y} \cap \hat{\mathbf{y}}|}{|\mathbf{y}| + |\mathbf{\hat{y}|}}, \quad 
      \text{where} \quad |\mathbf{y} \cap \hat{\mathbf{y}}| = \sum\limits_{i=1}^{h}\sum\limits_{j=1}^{w}y_{i,j} \times\hat{y}_{i,j}, \quad \text{and} \quad |\mathbf{y}|= \sum\limits_{i=1}^{h}\sum\limits_{j=1}^{w}y_{i,j}.
\end{align}

\paragraph{Intersection over Union.}
Intersection over Union (IoU) is a similar widely used segmentation metric defined as follows:
\begin{align}
\text{IoU}(\mathbf{y},\hat{\mathbf{y}}) = \frac{|\mathbf{y} \cap \hat{\mathbf{y}}|}{|\mathbf{y} \cup \hat{\mathbf{y}}|},
\quad \text{where} \quad |\mathbf{y} \cup \hat{\mathbf{y}}| = |\mathbf{y}| + |\hat{\mathbf{y}}| - |\mathbf{y} \cap \hat{\mathbf{y}}| .
\end{align}
Compared with DSC, IoU is generally a less forgiving metric since it does not weight the overlap as strongly, thereby penalising mismatches more strongly.

\paragraph{Generalised Energy Distance.}
We use Generalised Energy Distance (GED) to compare the quality of samples from the model with the ground truth labels. Independent samples $\mathbf{\hat{y}},\mathbf{\hat{y}'} \overset{\text{iid}}{\sim} p_{\text{model}}$ are drawn from the predictive model distribution $p_{\text{model}}$ whereas the multiple ground truth annotations $\mathbf{y},\mathbf{y'} \sim p_{\text{data}}$ are from the data distribution $p_{\text{data}}$, then:
\begin{align}
D^2_{\text{GED}}(p_{\text{data}}, p_{\text{model}}) = 2\, \mathbb{E}_{\mathbf{y} \sim p_{\text{data}}, \hat{\mathbf{y}} \sim p_{\text{model}}} [d(\mathbf{y}, \hat{\mathbf{y}})] - \mathbb{E}_{\hat{\mathbf{y}}, \hat{\mathbf{y}}' \sim p_{\text{model}}} [d(\hat{\mathbf{y}}, \hat{\mathbf{y}}')] - \mathbb{E}_{\mathbf{y}, \mathbf{y}' \sim p_{\text{data}}} [d(\mathbf{y}, \mathbf{y}')].
\end{align}
We follow \citet{kohl2018probabilistic,monteiro2020stochastic} by using the distance function $d(\mathbf{y},\mathbf{\hat{y}}) = 1 - \text{IoU}(\mathbf{y},\mathbf{\hat{y}})$, which is a metric, and implies $D^2_{\text{GED}}$ is also a metric. Lower GED indicates better alignment between the predictive and ground truth distributions. The sample diversity $\mathbb{E}_{\hat{\mathbf{y}}, \hat{\mathbf{y}}' \sim p_{\text{model}}} [d(\hat{\mathbf{y}}, \hat{\mathbf{y}}')]$ is also a quantity of interest, as it measures the average distance between pairs of samples from the predictive distribution, and provides a measure of variability in our samples. We note that diversity is only contextually relevant, as high diversity alone can be trivially achieved with random noise as a model.

\paragraph{Hungarian-matched IoU.}
We also use Hungarian-matched IoU (HM-IoU) to compare samples between the ground truth and predictive distributions. HM-IoU uses the Hungarian algorithm to find an optimal assignment between the two sets of samples: $\{\mathbf{y}_1, \mathbf{y}_2, \dots, \mathbf{y}_N\} \sim p_{\text{data}}$ and $\{\hat{\mathbf{y}}_1, \hat{\mathbf{y}}_2, \dots, \mathbf{\hat{y}}_M\}\overset{\text{iid}}{\sim} p_{\text{model}}$, then using $1-\text{IoU}$ as the cost matrix for solving a linear sum assignment problem~\cite{kohl2019hierarchical}. HM-IoU can be viewed as more robust across sets of ground truth and predictive samples compared to the GED, which can suffer from inflated scores if the predictive samples are very diverse.
\section{Implementation Details}
\label{app: implementation}
\subsection{Training Setup \& Hyperparameters}
In this section, we provide the setup and hyperparameters we used to train our Flow-SSN models on the medical datasets. Our implementation is based in \texttt{PyTorch}~\cite{paszke2019pytorch}. As shown in Table~\ref{tab:hyperparams}, we use a UNet~\cite{ronneberger2015u,dhariwal2021diffusion} to parameterise the base distribution of our Flow-SSN (i.e. Prior Network). To parameterise the flow transformation itself (i.e. Flow Network), we use an autoregressive Transformer for the discrete-time Flow-SSN variant and a UNet for the continuous-time variant. For data augmentation on LIDC-IDRI, we used random 90 degree rotations and vertical/horizontal flips with 0.5 probability. For data augmentation on REFUGE-MultiRater, we used random vertical flips with 0.5 probability; random rotations in the range of [-20, 20] degrees, random resized cropping to 256x256 with scale [0.9, 1.1], and applied color jitter of 0.3 to the images.

The remaining hyperparameters we used are largely equal for both datasets, as reasonable initial values performed well enough; we did not perform extensive hyperparameter tuning for each dataset. For the continuous-time Flow-SSN, 8 ODE solving steps (Euler method in \texttt{torchdiffeq}~\cite{torchdiffeq}) were used on the validation set throughout training for model selection. In all cases, the best checkpoint was selected based on the lowest GED achieved on the validation set during training. The final model artefact we use for evaluation is an exponential moving average (EMA) of the model parameters. 
\begin{table}[h] 
\caption{Training hyperparameters used across all experiments.}\label{tab:hyperparams}
\centering 
\begin{tabular}{lcccc}
\toprule
 & \multicolumn{2}{c}{\textbf{LIDC-IDRI}} & \multicolumn{2}{c}{\textbf{REFUGE-MultiRater}} 
 \\ 
 \textsc{Config} & Flow-SSN$_{\Delta}$ & Flow-SSN$_{\infty}$ & Flow-SSN$_{\Delta}$ & Flow-SSN$_{\infty}$ \\ \midrule
 {Prior Network} & UNet & UNet & UNet & UNet \\
 {Flow Network} & Transformer & UNet &
 Transformer & UNet \\
 {Optimiser} & AdamW & AdamW & AdamW & AdamW \\
  {Batch Size} & 16 & 16 & 16 & 16 \\
 Learning Rate & $10^{-4}$ & $10^{-4}$ & $10^{-4}$ & $10^{-4}$ \\
 {LR Warmup} & Linear 2K & Linear 2K & Linear 1K & Linear 1K \\
  Weight Decay & $10^{-4}$ & $10^{-4}$ & $10^{-4}$ & $10^{-4}$ \\
EMA Rate & 0.9999 & 0.9999 & 0.999 & 0.999 \\
  Max Epochs & 1001 & 1001 & 1001 & 1001
 \\ 
   Eval Freq. & 16 & 16 & 50 & 50
 \\ 
MC Samples (train) & 16 & 1 & 32 & 1 \\
MC Samples (eval) & 16 & 16 & 16 & 16 \\
 Prior Dist. & Gaussian & Gaussian & Gaussian & Gaussian
 \\ 
 \bottomrule
\end{tabular}
\end{table}

\subsection{UNet Architecture}
As mentioned in the main text, we reimplemented a streamlined version of~\citet{dhariwal2021diffusion}'s UNet, which uses fewer attention layers. Recall that LIDC-IDRI images are grayscale, whereas REFUGE images are RGB. To parameterise a Flow-SSN prior, the UNet outputs a mean and variance per output pixel, representing the `initial guess' distribution as a diagonal Gaussian, to be later refined by the choice of flow. The prior's mean can be optionally initialised with a pre-trained network and then fine-tuned alongside the flow network. In our case, we train everything end-to-end for simplicity and find that fixing the prior variance to, e.g. 1, can also help stabilise training in some instances. Table~\ref{tab:hyperparams_unet} shows the remaining details, including input and output shapes for both the prior and flow networks.
\begin{table}[!h] 
\caption{UNet hyperparameters used for parameterising both the base distribution (Prior) and the flows in our Flow-SSN models. We use ($\Delta$) and ($\infty$) to denote relation to the discrete and continuous-time version of the associated Flow-SSN.}\label{tab:hyperparams_unet}
\centering 
\begin{tabular}{lcccc}
\toprule
 & \multicolumn{2}{c}{\textbf{LIDC-IDRI}} & \multicolumn{2}{c}{\textbf{REFUGE-MultiRater}} 
 \\ 
 \textsc{Config} & Prior (${\Delta,\infty}$) & Flow (${\infty}$) & Prior (${\Delta,\infty}$) & Flow (${\infty}$) 
 \\ 
 \midrule
 Input Shape & (1, 128, 128) & (2, 128, 128) & (3, 256, 256) & (2, 256, 256)
 \\
 Model Channels & 32 & 16 & 32 & 32 \\
Output Channels & 4 & 2 & 4 & 2 \\
Residual Blocks & 1 & 1 & 2 & 1 \\
Dropout & 0.1 & 0.1 & 0.1 & 0.1 \\
 Channel Multipliers & [1, 2, 4, 8] & [1, 1, 1] & [1, 2, 2, 4, 6] & [1, 1, 1, 1] 
 \\ 
 Attention Resolution & [16] & [16] & [16] & [16] \\
 Num. Heads & 1 & 1 & 1 & 1 \\
 Head Channels & 64 & 16 & 64 & 32
 \\
 \#Parameters & 14.4M & 150K & 14.6M & 787K
 \\
 \bottomrule
\end{tabular}
\end{table}

\subsection{Autoregressive Transformer Architecture}
To parameterise the discrete-time autoregressive Flow-SSN, we require a lightweight autoregressive model. The challenge is that our flow lives in pixel-space, so autoregressively predicting each pixel can become computationally expensive for large datasets. To overcome this obstacle, we propose two things. The first is that we use an IAF~\cite{kingma2016improved} defined in \textit{in pixel-space} and train under the expected likelihood objective in Eq.~\eqref{eq:discrete_xlike}, which avoids pixel-wise sequential likelihood evaluation and enables fast, one-pass sampling at inference time. Recall that MAFs~\cite{papamakarios2017masked} are equally fast to train as likelihood evaluation can be parallelised, but they still require sequential autoregressive sampling at inference time. The second is that we use an autoregressive Transformer to parameterise the flow, with image strips/patches of, e.g., size (1,8) or (8,8) pixels to reduce memory requirements. To reconstruct the patches back to the output size, we simply use a transposed convolution. Grouping pixels this way induces a block covariance structure in pixel space, but as long as the strips/patches are small relative to the full image size, the maximum attainable covariance rank remains high (i.e. in the hundreds/thousands for real images). Furthermore, since Flow-SSNs model pixel covariance in logit space and transform the learned distribution by a non-linear transfer function (softmax), there is a sublinear increase in rank we can benefit from (c.f. theoretical results in Appendix A).

In summary, and perhaps surprisingly, we find that the above combination of design choices works well provided we use enough MC samples during training, e.g. $\geq8$, otherwise, the model can collapse to a near-deterministic state (c.f. Fig.~\ref{fig:iaf_mc_refuge}). It is worth noting that we briefly experimented with a shallow PixelCNN~\cite{van2016pixel,van2016pixel,salimans2017pixelcnn} to parameterise the flow instead of an autoregressive Transformer, but did not have as much success. As shown in Table~\ref{tab:hyperparams_transformer}, the final architecture setup for a discrete-time autoregressive Flow-SSN uses just 1 autoregressive flow transformation parameterised by a single Transformer layer! As proven in Appendix A, a single autoregressive transformation is sufficient to transform the diagonal Gaussian prior into a highly expressive full covariance. We confirm this works well in practice on both toy and real medical imaging datasets.
\begin{table}[h] 
\caption{Autoregressive Transformer hyperparameters used for our discrete-time autoregressive Flow-SSNs.}\label{tab:hyperparams_transformer}
\centering 
\begin{tabular}{lcc}
\toprule
 & \textbf{LIDC-IDRI} & \textbf{REFUGE-MultiRater}
 \\ 
 \textsc{Config} & Flow ($\Delta$) & Flow (${\Delta}$)
 \\
 \midrule
 Input Shape & (2, 128, 128) & (2, 256, 256)
 \\
Num. Flows & 1 & 1
 \\
 Flow Type & IAF & IAF
 \\
   Output Channels & 4 & 4
 \\
    Embed Dim & 64 & 128
 \\
     MLP Width & 256 & 512
 \\
   Patch Size & (1, 8) & (8, 8)
 \\
 Patchify & Conv & Conv
 \\
 Unpatchify & ConvTranspose & ConvTranspose
 \\
     Num. Blocks & 1 & 2
 \\
     Num. Heads & 1 & 1
 \\
    Pos Embed Init & $\mathcal{N}(0, 0.02)$ & $\mathcal{N}(0, 0.02)$
\\
    Dropout & 0.1 & 0.1
 \\
     Activation & GELU & GELU
 \\
     \#Parameters & 345K & 0.92M
 \\
 \bottomrule
\end{tabular}
\end{table}

\newpage
\section{Extra Results}
\label{app: extra results}
\subsection{Sampling Efficiency \& Additional Baselines}
As reported in Table~\ref{tab:sota-comp2}, our method is ${\approx}10\times$ more efficient than CCDM~\citep{zbinden2023stochastic}, thanks to most of the model parameters being in the flow's \textit{prior} (i.e. base/source distribution), which only requires a single forward pass to start the sampling chain. Afterwards, only the \textit{flow} network is needed to solve the ODE, and it has only 150K parameters\footnote{For 1-step we use a discrete-time inverse autoregressive Flow-SSN}. The advantage of our model translates to any other diffusion-based segmentation model that dedicates all its model capacity to learning the score/velocity field~\cite{amit2021segdiff,rahman2023ambiguous,zbinden2023stochastic,wu2024medsegdiff}. That said, we expect that using large models for both the prior \textit{and} the flow would improve performance even further; though we argue that the latter may not be necessary if the prior is expressive enough, e.g. a foundation model. Table~\ref{tab:sota-comp} reports further baseline comparisons against recent SOTA methods; Flow-SSN outperforms all previous methods.
\begin{table}[!h]
\centering
\caption{Comparing Flow-SSN sampling efficiency against CCDM (see Figure 5 in \cite{zbinden2023stochastic}).}
\begin{tabular}{l|rrc|ccc}
\toprule
& \multicolumn{3}{c|}{CCDM~\citep{zbinden2023stochastic}} & \multicolumn{3}{c}{Flow-SSN}
\\[2pt]
Steps & $D^2_{\text{GED}}(16) \downarrow$ & HM-IoU $\uparrow$ & FLOPS $\downarrow$ & $D^2_{\text{GED}}(16) \downarrow$ & HM-IoU $\uparrow$ & FLOPS $\downarrow$
\\ 
\midrule
1 & \multicolumn{1}{c}{-} & \multicolumn{1}{c}{-} & - & 0.240\scriptsize{$\pm$.002} & {0.879}\scriptsize{$\pm$.000} & 12G              
\\
50 & $\simeq$ 0.34{\scriptsize $\pm$N/A} & $\simeq$ 0.55{\scriptsize $\pm$N/A} & \cellcolor{red!10}365G & {0.210}\scriptsize{$\pm$.002} & {0.873}\scriptsize{$\pm$.000} & \cellcolor{green!10}\textbf{51G}
\\
250 & 0.212\scriptsize{$\pm$.002} & 0.623\scriptsize{$\pm$.002} & \cellcolor{red!10}1.83T & {0.207}\scriptsize{$\pm$.000} & {0.873}\scriptsize{$\pm$.001} &  \cellcolor{green!10}\textbf{207G}
\\
\bottomrule
\end{tabular}
\label{tab:sota-comp2}
\end{table}
\begin{table}[!h]
\centering
\caption{Additional comparisons of Flow-SSN against recent SOTA methods on LIDC-IDRI.}
\begin{tabular}{l|rccr}
\toprule
\centering
\textsc{Method} & Pub. Venue & $D^2_{\text{GED}} (16) \downarrow$ & HM-IoU $\uparrow$ & \#Param \\ 
\midrule
Tyche~\citep{rakic2024tyche}                 & CVPR'24                             & 0.400\scriptsize{$\pm$.010}                                         & -                                     & 1.7M\\
CCDM~\citep{zbinden2023stochastic}                  & ICCV'23                             & 0.212\scriptsize{$\pm$.002}                                       & 0.623\scriptsize{$\pm$.002}                      & 9M\\
MoSE~\citep{gao2023modeling}                 & ICLR'23                             & 0.218\scriptsize{$\pm$.003}                                       & 0.624\scriptsize{$\pm$.004}                      & 42M \\ 
CIMD~\citep{rahman2023ambiguous}                  & CVPR'23                             & 0.234\scriptsize{$\pm$.005}                                       & 0.587\scriptsize{$\pm$.001}                      & 24M\\ 
\midrule
{Flow-SSN}$_\infty$              &                 ICCV'25                      & \textbf{0.207}\scriptsize{$\pm$.000}                                       & \textbf{0.873}\scriptsize{$\pm$.001}                      & 14M\\
\bottomrule
\end{tabular}
\label{tab:sota-comp}
\end{table}
\subsection{Ablation Study: Increasing the Assumed Rank}
\begin{figure*}[!h]
    \centering
    \begin{subfigure}{.41\textwidth}
        \centering
        \includegraphics[width=\textwidth]{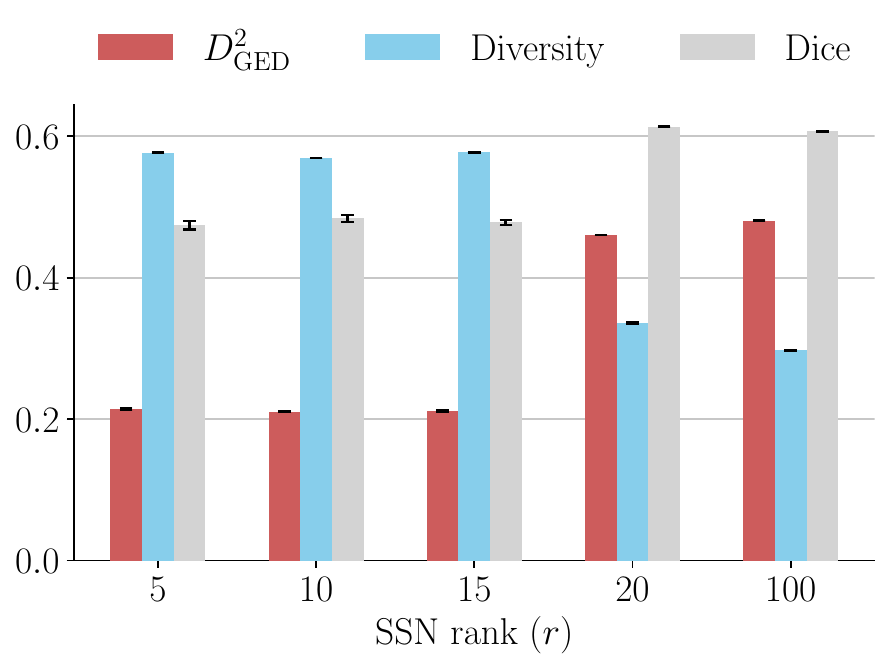}
    \end{subfigure}
    \hspace{20pt}
    \begin{subfigure}{.51\textwidth}
        \centering
        \includegraphics[width=\textwidth]{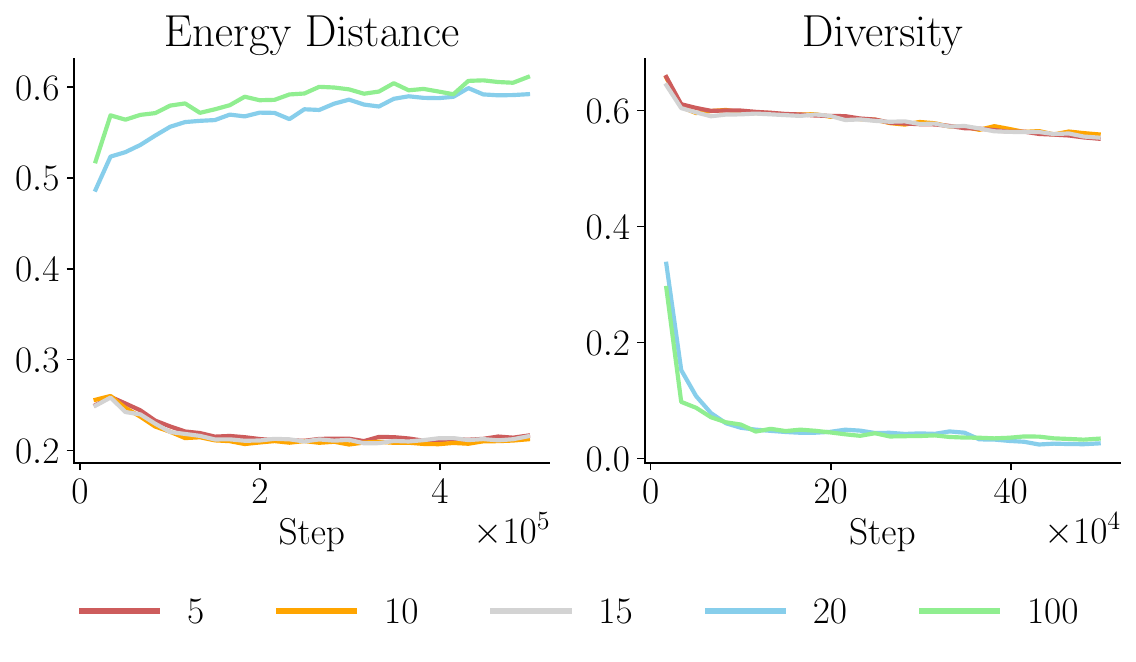}
    \end{subfigure}
    \caption{\textbf{Scalability ablation study of SSNs on LIDC in terms of the assumed rank}. The results show that mild increases in the assumed rank can cause SSNs to collapse into a near-deterministic state. This warrants a new approach for estimating high-rank covariances.}
    \label{fig:ssn_scale}
\end{figure*}

We first assess the scalability of standard SSNs by ablating an increase in the assumed rank using a replication of~\citet{monteiro2020stochastic}'s setup. We observe (Figure~\ref{fig:ssn_scale}) that SSNs tend to collapse to deterministic models even under mild increases in the assumed rank, which shows that a different approach is needed to capture higher-order interactions between pixels.
\subsection{Ablation Study: Number of Monte Carlo Samples}
\begin{figure*}[!h]
    \centering
    \includegraphics[width=.925\textwidth]{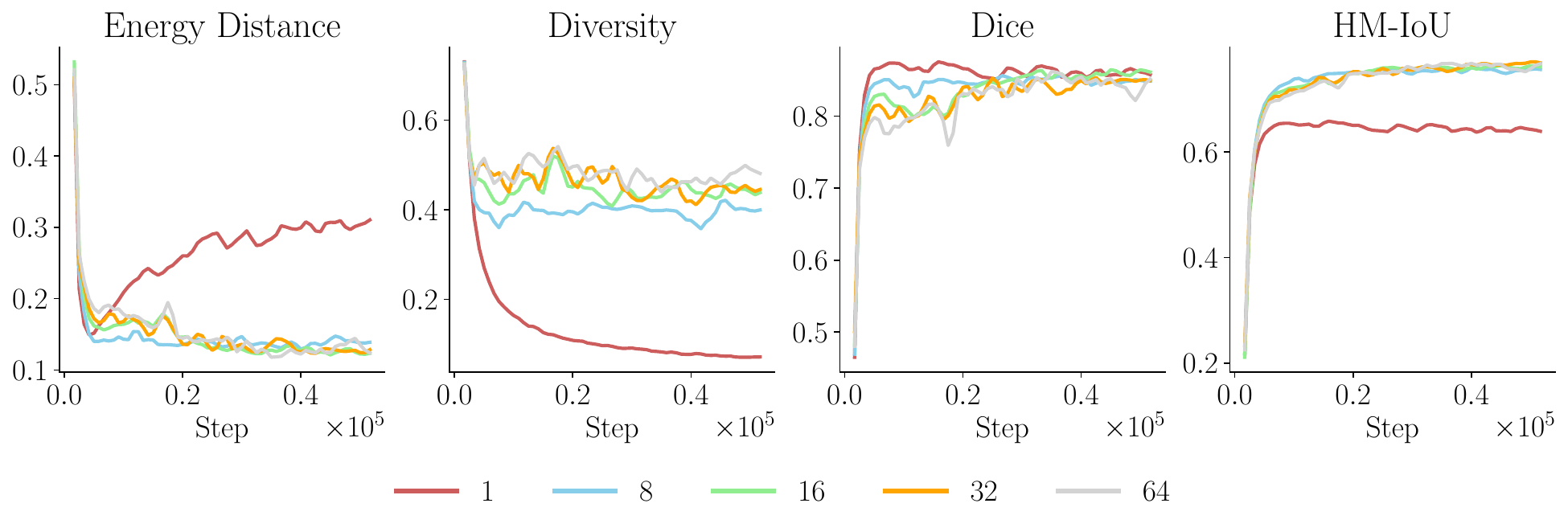}
    \caption{\textbf{Ablation analysis of the number of Monte Carlo (MC) samples needed for training}. We look at $\{1, 8, 16, 32, 64\}$ used for training a discrete-time autoregressive Flow-SSN with the objective in Equation~\eqref{eq:discrete_xlike}. The above results were obtained using the REFUGE-MultiRater dataset, and performance on the validation set is shown throughout training.}
    \label{fig:iaf_mc_refuge}
\end{figure*}

As shown in Figure~\ref{fig:iaf_mc_refuge}, we find that multiple MC samples are needed to properly learn the underlying distribution over outputs when using a discrete-time inverse autoregressive Flow-SSN. With only 1 MC sample, the model enters the near-deterministic regime, and increasing the number of MC samples provides diminishing returns in terms of improved performance. We consider further improving the training stability of IAFs at scale, possibly borrowing tricks from modern MAFs and coupling flows~\citep{tschannen2025jetformer,zhai2025normalizing,gu2025starflow}, to be fertile ground for future work. The payoff in efficiency is significant as sampling becomes parallelizable.
\newpage
\subsection{Qualitative Results: LIDC-IDRI}
\begin{figure*}[!h]
    \centering
    \includegraphics[clip, trim={0 0 0 0}, width=\textwidth]{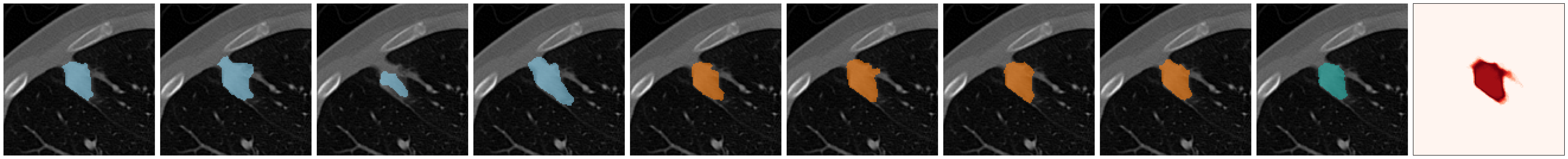}
    \\
    \includegraphics[clip, trim={0 0 0 0}, width=\textwidth]{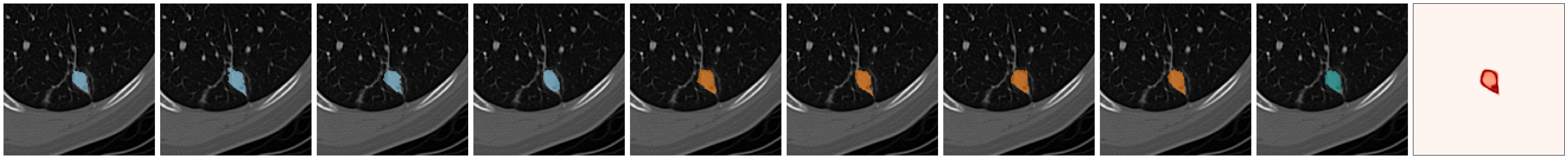}
    \\
    \includegraphics[clip, trim={0 0 0 0}, width=\textwidth]{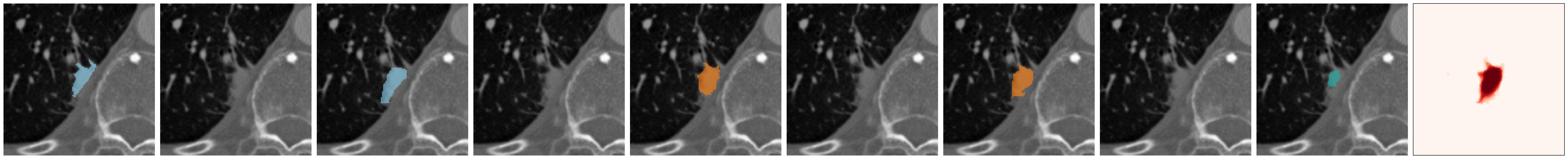} 
    \\
    \includegraphics[clip, trim={0 0 0 0}, width=\textwidth]{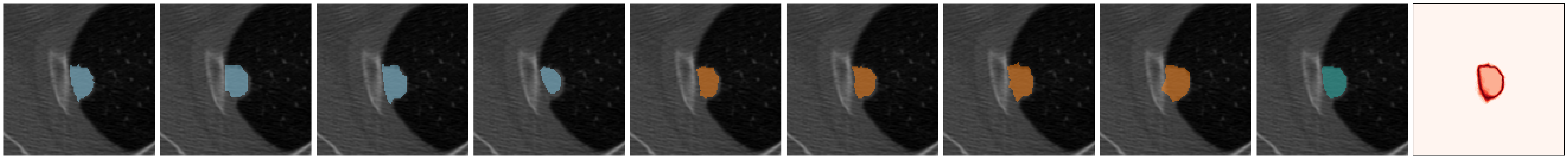}
    \\
    \includegraphics[clip, trim={0 0 0 0}, width=\textwidth]{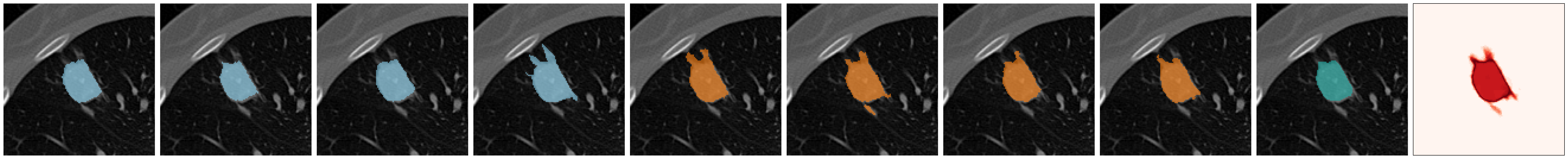}
    \\
    \includegraphics[clip, trim={0 0 0 0}, width=\textwidth]{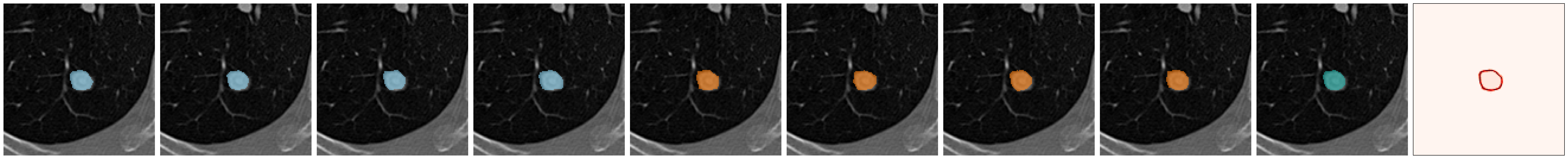}
    \caption{\textbf{Extra qualitative results on LIDC-IDRI using our continuous-time Flow-SSN model}. (\textit{Cols. 1-4}) Multiple ground truth segmentations from experts; (\textit{Cols. 5-8}) Non-cherry-picked random samples from our model; (\textit{Cols. 9, 10}) The mean prediction and per-pixel uncertainty map. In all cases, 100 MC samples and 50 ODE solving steps (Euler method) were used for evaluation.}
    \label{fig:lidc_samples_extra}
\end{figure*}
\begin{figure*}[!h]
    \centering
    \includegraphics[width=.999\textwidth]{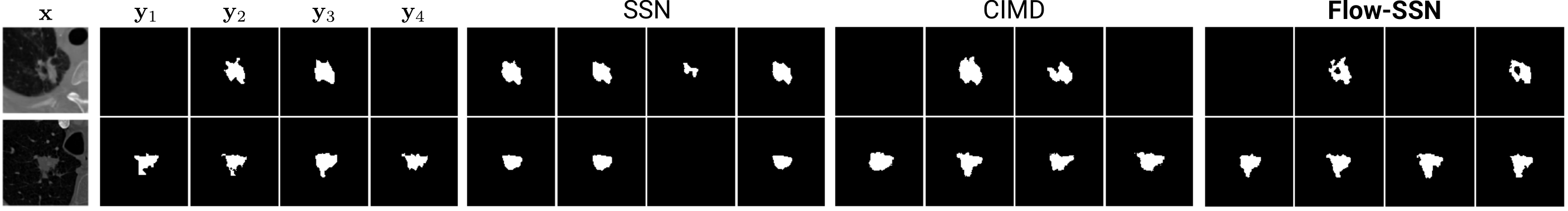}
    \caption{\textbf{Qualitative comparison of Flow-SSN against previous methods on LIDC-IDRI}. We used the same images as CIMD~\citep{rahman2023ambiguous} for fair comparisons. (\textit{Col. 1}) Input images for segmentation; (\textit{Cols. 2-5}) Ground truth annotations from multiple experts (i.e. four total in this case $\{\mathbf{y}_1, \mathbf{y}_2, \mathbf{y}_3, \mathbf{y}_4\}$); (\textit{Cols. 6-9}) Random samples from our reproduced SSN~\cite{monteiro2020stochastic} model; (\textit{Cols. 10-13}) Random samples taken from CIMD~\citep{rahman2023ambiguous}, a diffusion-based segmentation model; (\textit{Cols. 14-17}) Random samples from our (continuous-time) Flow-SSN model.
    }
    \label{fig:lidc_samples_comp}
\end{figure*}
\newpage
\subsection{Qualitative Results: REFUGE-MultiRater}
\begin{figure*}[!h]
    \centering
    \includegraphics[clip, trim={0 0 0 0}, width=\textwidth]{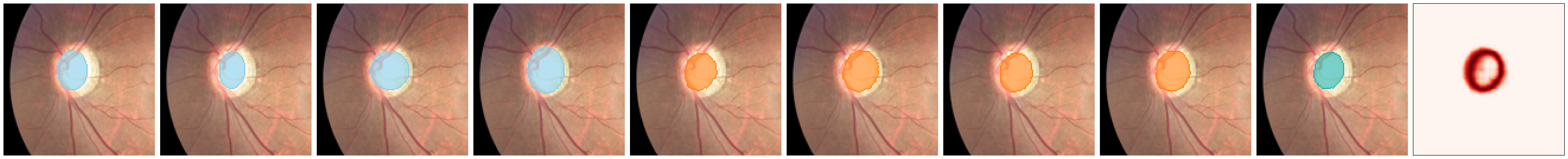}
    \\
    \includegraphics[clip, trim={0 0 0 0}, width=\textwidth]{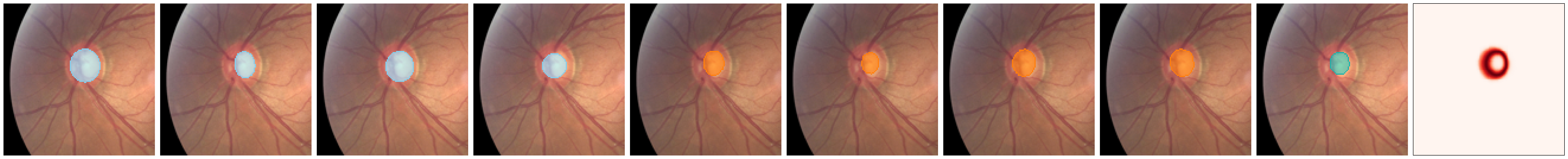}
    \\
    \includegraphics[clip, trim={0 0 0 0}, width=\textwidth]{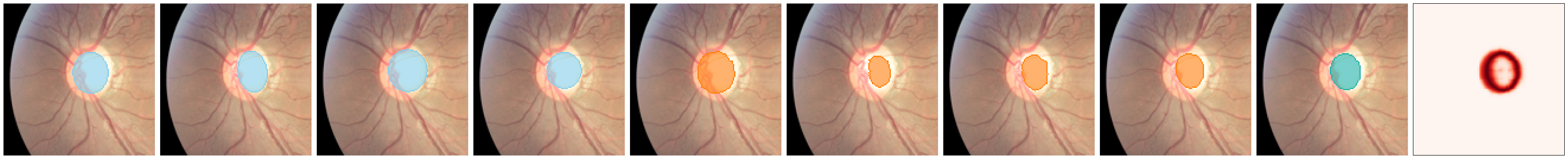}
    \\
    \includegraphics[clip, trim={0 0 0 0}, width=\textwidth]{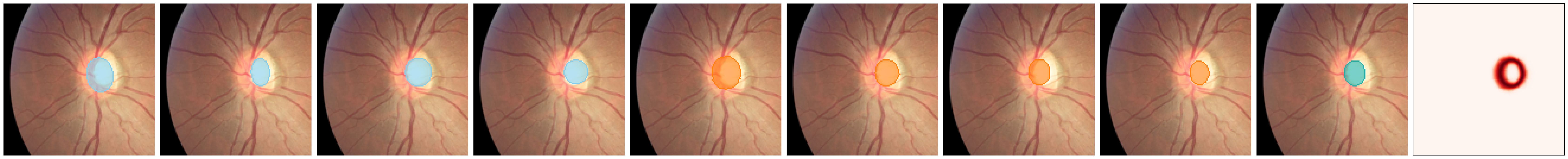}
    \\
    \includegraphics[clip, trim={0 0 0 0}, width=\textwidth]{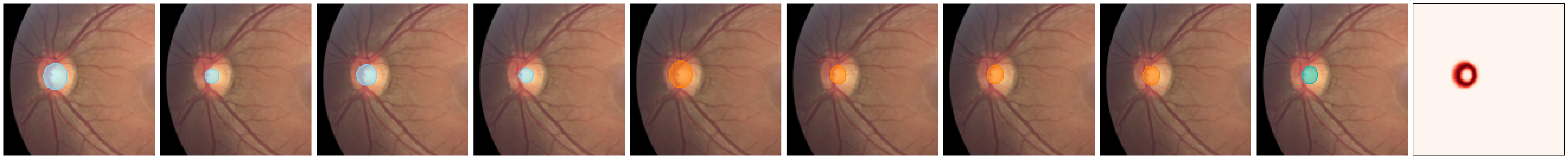}
    \\
    \includegraphics[clip, trim={0 0 0 0}, width=\textwidth]{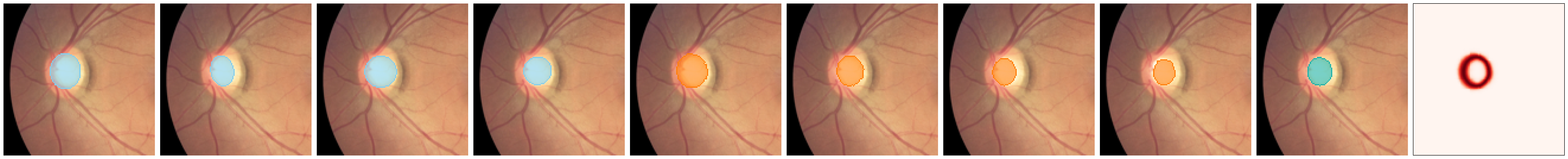}
    \\
    \includegraphics[clip, trim={0 0 0 0}, width=\textwidth]{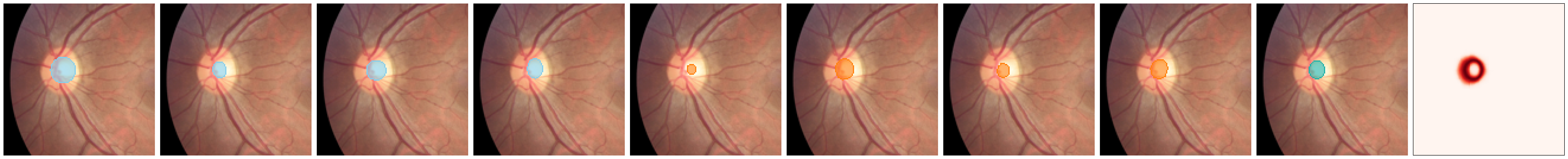}
    \caption{\textbf{Extra qualitative results on REFUGE-MultiRater using our discrete-time autoregressive Flow-SSN model}. (\textit{Cols. 1-4}) Multiple ground truth segmentations from experts; in each case, four segmentations were randomly chosen out of the seven available in total; (\textit{Cols. 5-8}) Non-cherry-picked random samples from our discrete-time autoregressive Flow-SSN model; (\textit{Cols. 9, 10}) The mean prediction and per-pixel uncertainty map. In all cases, 512 Monte Carlo samples were used for performing the evaluation.}
    \label{fig:refuge_samples_extra}
\end{figure*}

\end{document}